\documentclass[lettersize,journal]{IEEEtran}
\usepackage{amsmath,amsfonts}
\usepackage{algorithmic}
\usepackage{array}
\usepackage{textcomp}
\usepackage{stfloats}
\usepackage{url}
\usepackage{verbatim}
\usepackage{graphicx}
\PassOptionsToPackage{table}{xcolor}
\newcommand{\ie}{\textit{i.e.}}
\newcommand{\eg}{\textit{e.g.}}

\usepackage{multirow}
\usepackage{makecell}
\usepackage{tikz}
\usetikzlibrary{shapes.geometric}
\usetikzlibrary{fit}
\usetikzlibrary{decorations.pathreplacing}
\usetikzlibrary{positioning, shapes, calc, arrows, arrows.meta, backgrounds, shadows, matrix}
\usepackage{arydshln} 
\usepackage{booktabs}
\usepackage{pifont}
\usepackage{pgfplots}
\pgfplotsset{compat=1.17}
\usepackage{bm} 
\usepackage{fontawesome5} 
\usepackage{adjustbox}
\definecolor{lightgray}{gray}{0.95}
\usepackage{subcaption}
\usepackage{orcidlink}
\def\BibTeX{{\rm B\kern-.05em{\sc i\kern-.025em b}\kern-.08em
    T\kern-.1667em\lower.7ex\hbox{E}\kern-.125emX}}
\usepackage{balance}
\begin{document} 
\title{From Data to Modeling: Fully Open-vocabulary Scene Graph Generation}
\author{Zuyao Chen \orcidlink{0000-0002-7344-1101}, \and
Jinlin Wu \orcidlink{0000-0001-7877-5728},  \and
Zhen Lei \orcidlink{0000-0002-0791-189X}, \emph{Fellow}, IEEE, \and 
and Chang Wen Chen \orcidlink{0000-0002-6720-234X}, \emph{Fellow}, IEEE \thanks{Zuyao Chen and Chang Wen Chen are with the Department of Computing, The Hong Kong Polytechnic University, Hong Kong (e-mail: zuyao.chen@connect.polyu.hk; changwen.chen@polyu.edu.hk). 

Jinlin Wu is with CAIR, HKISI-CAS, Hong Kong SAR and MAIS, Institute of Automation, Chinese Academy of Sciences, China (e-mail: jinlin.wu@cair-cas.org.hk). 

Zhen Lei is with CAIR, HKISI-CAS, Hong Kong SAR, MAIS, Institute of Automation, Chinese Academy of Sciences, China, and School of Artificial Intelligence, University of Chinese Academy of Sciences, China (e-mail: zhen.lei@ia.ac.cn).

Our code is available at \url{https://github.com/gpt4vision/OvSGTR}. 
}
}

\markboth{Journal of \LaTeX\ Class Files,~Vol.~18, No.~9, September~2020}%
{How to Use the IEEEtran \LaTeX \ Templates}

\maketitle
\begin{abstract}
We present OvSGTR, a novel transformer-based framework for fully open-vocabulary scene graph generation that overcomes the limitations of traditional closed-set models. 
Conventional methods restrict both object and relationship recognition to a fixed vocabulary, hindering their applicability to real-world scenarios where novel concepts frequently emerge. 
In contrast, our approach jointly predicts objects (nodes) and their inter-relationships (edges) beyond predefined categories. 
OvSGTR leverages a DETR-like architecture featuring a frozen image backbone and text encoder to extract high-quality visual and semantic features, which are then fused via a transformer decoder for end-to-end scene graph prediction. 
To enrich the model's understanding of complex visual relations, we propose a relation-aware pre-training strategy that synthesizes scene graph annotations in a weakly supervised manner. 
Specifically, we investigate three pipelines—scene parser-based, LLM-based, and multimodal LLM-based—to generate transferable supervision signals with minimal manual annotation. 
Furthermore, we address the common issue of catastrophic forgetting in open-vocabulary settings by incorporating a visual-concept retention mechanism coupled with a knowledge distillation strategy, ensuring that the model retains rich semantic cues during fine-tuning. 
Extensive experiments on the VG150 benchmark demonstrate that OvSGTR achieves state-of-the-art performance across multiple settings, including closed-set, open-vocabulary object detection-based, relation-based, and fully open-vocabulary scenarios. 
Our results highlight the promise of large-scale relation-aware pre-training and transformer architectures for advancing scene graph generation towards more generalized and reliable visual understanding.
\end{abstract}


\section{Introduction}
\definecolor{col1}{RGB}{0,0,255} 
\definecolor{col2}{RGB}{0,255,255} 
\definecolor{col3}{RGB}{255,0,0} 
\definecolor{col4}{RGB}{255,0,0} 
\definecolor{colorsgg}{RGB}{217,217,217} 
\definecolor{colorArrow}{RGB}{64,64,64} 

\definecolor{color_man}{RGB}{151, 178, 228}
\definecolor{color_sk}{RGB}{175, 248, 204}
\definecolor{color_helmet}{RGB}{102, 212, 119}
\definecolor{color_wheel}{RGB}{152, 114, 178}
\definecolor{color_shirt}{RGB}{218, 176, 225}
\definecolor{color_pant}{RGB}{114, 129, 157}
\definecolor{color_flow}{RGB}{173, 216, 230}
\newcommand{\scaleF}{0.65}
\tikzset{
  myarrow/.style={
    ->,                    
   -{Latex[round]},             
    thin, 
    draw=colorArrow,       
    line width=1pt,      
    text=black,            
    shorten >=0.4pt,         
    shorten <=0pt,           
  },
sggnode/.style={rectangle, rounded corners=2pt, draw, thick, fill=colorsgg, text width=1.2cm, align=center, drop shadow}
}
\begin{figure*}[t]
\centering 
\definecolor{color_man0}{RGB}{142, 181, 113}
\definecolor{color_pant0}{RGB}{138, 236, 158}
\definecolor{color_h0}{RGB}{225, 200, 152}
\begin{minipage}{\textwidth}
\centering
    \begin{tikzpicture}[node distance=8mm, auto, scale=\scaleF, every node/.style={scale=\scaleF},
]
        \node (img1) [] {\includegraphics[width=0.7in]{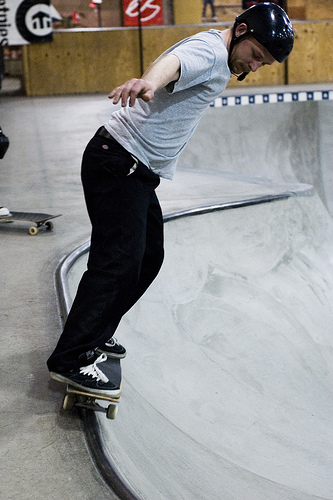}};
        \node (box1) [right=of img1, xshift=8mm, sggnode] {SGG Model}; 
        \node (user) [above=of box1,  xshift=-6mm] {
            \includegraphics[width=0.3in]{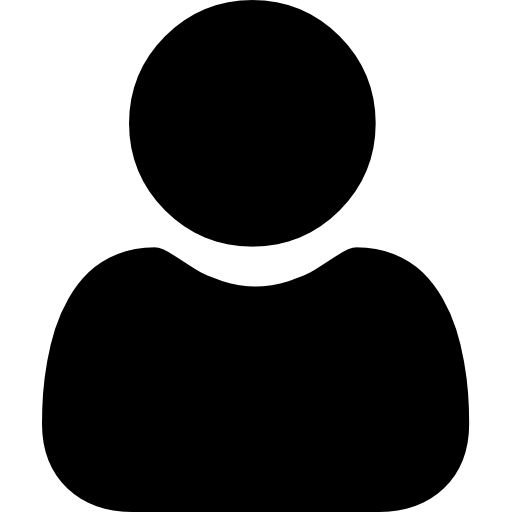}
        }; 
        \node [right of=user, xshift=4mm] {User};

        \node [above =of box1, yshift=-8mm, xshift=4.4mm] {Fixed Vocabulary (Object \& Relationship)};
        
        \draw[myarrow] (img1) to (box1);
        \draw[myarrow] ([xshift=6mm]user.south) to (box1); 
        \draw[myarrow] (box1) to ([xshift=12mm]box1.east);

         \node[draw, circle, fill=color_man, right =of box1, xshift=25mm] (man) {};
            \node[right of=man, xshift=-2mm] {man}; 
            \node[draw, circle, below right =of man, fill=color_pant] (pant) {};
            \node[right of=pant, xshift=-2mm] {pant}; 
            \node[draw, circle, below left=of man, fill=color_helmet] (helmet) {};
             \node[left of=helmet, xshift=0mm] {helmet};

            \draw[->, >=stealth] (man) -- node[midway, sloped, anchor=center, above] {in} (pant);
            \draw[->, >=stealth] (helmet) -- node[midway, sloped, anchor=center, above] {on} (man);

            \node[star, star points=5, star point ratio=2.25, fill=col3!80, draw, inner sep=0pt, minimum size=8pt, left=of man, xshift=-7mm] {};
            
    \end{tikzpicture}  
    \subcaption{\emph{Closed-set SGG}.}
\end{minipage}
\vfill 
    \begin{minipage}[c]{\textwidth}
   \vspace{-2mm}
    \centering
    \begin{tikzpicture}[node distance=8mm, auto, scale=\scaleF, every node/.style={scale=\scaleF}]
        \node (A) at (0, 0) {};
        \node (B) at (-1, -1.) {};
        \node (A1) at (3, 0) {};
        \node (C) at (4, -1) {};
        
        \draw[line width=3pt, -{Triangle[length=3mm, width=3mm]}, color=color_flow] (A) to (B);
        \draw[line width=3pt, -{Triangle[length=3mm, width=3mm]}, color=color_flow] (A1) to (C);
    \end{tikzpicture}
    \end{minipage}
    \vfill 
\begin{minipage}{0.49\textwidth} 
\centering
    \begin{tikzpicture}[node distance=8mm, auto, scale=\scaleF, every node/.style={scale=\scaleF}]
    
        \node (img1) [] {\includegraphics[width=0.7in]{pics/2315830.jpg}};
        \node (box1) [right=of img1, xshift=-5mm, sggnode] {SGG Model}; 
        \node (cap1) [above =of img1, yshift=-12mm]{\emph{a man is on a \colorbox{color_sk}{skateboard}}}; 
        \draw[->, dashed, rounded corners=3pt] (cap1.east) --
        ([xshift=4.1mm]cap1.east) -- (box1.north); 
        \node [right=of cap1, xshift=-11mm, yshift=3mm] {Novel Objects}; 
        
        \node[draw, circle, fill=color_man, right=of box1, xshift=15mm] (man) {};
            \node[above of=man, yshift=-5mm] {man}; 
            \node[draw, circle, below right =of man, fill=color_pant] (pant) {};
            \node[right of=pant, xshift=-2mm] {pant}; 
            \node[draw, circle, below left=of man, fill=color_helmet] (helmet) {};
             \node[left of=helmet, xshift=0mm] {helmet};
            \node [draw, circle, thick, densely dashed, right =of man, fill=color_sk, xshift=-4.5mm,  fill opacity=0.5] (skateboard) {};
            \node [right of=skateboard, xshift=2mm] {skateboard}; 
            \node [draw, circle, above =of skateboard, fill=color_wheel] (wheel) {};
            \node [right of=wheel, xshift=-2mm] {wheel}; 

            \draw[->, >=stealth] (man) -- node[midway, sloped, anchor=center, above] {in} (pant);
            \draw[->, >=stealth] (helmet) -- node[midway, sloped, anchor=center, above] {on} (man);
            \draw[->, >=stealth] (wheel) -- node[midway, sloped, anchor=center, above] {on} (skateboard);

            \node (star1) [star, star points=5, star point ratio=2.25, fill=col3!80, draw, inner sep=0pt, minimum size=8pt, left=of wheel, xshift=-18mm] {};
            \node[star, star points=5, star point ratio=2.25, fill=col3!80, draw, inner sep=0pt, minimum size=8pt, right of=star1, xshift=-5mm] {};

        \draw[myarrow] (img1) to (box1);
        \draw[myarrow] (box1) to ([xshift=5mm]box1.east);
    \end{tikzpicture}
    \subcaption{\emph{OvD-SGG}.}
\end{minipage} \hfill 
\begin{minipage}{0.5\textwidth}
\centering
        \begin{tikzpicture}[node distance=8mm, auto, scale=\scaleF, every node/.style={scale=\scaleF}]
            
            \node[draw, circle, fill=color_man] (man) {};
            \node[right of=man, xshift=-3mm, yshift=2mm] {man}; 
            \node[draw, circle, below right =of man, fill=color_pant] (pant) {};
            \node[right of=pant, xshift=-2mm] {pant}; 
            \node[draw, circle, below left=of man, fill=color_helmet] (helmet) {};
            \node[left of=helmet, xshift=0mm] {helmet};
            \node[draw, circle, above =of man, fill=color_shirt] (shirt) {};
            \node[right of=shirt, xshift=-2mm] {shirt}; 
 
            \draw[->, >=stealth] (man) -- node[midway, sloped, anchor=center, above] {in} (pant);
            \draw[->, thick, densely dashed, >=stealth] (man) -- node[midway, sloped, anchor=center, above] {\textcolor{pink}{wearing}} (shirt);
            \draw[->, >=stealth] (helmet) -- node[midway, sloped, anchor=center, above] {on} (man);

            \node [above=of shirt, yshift=-10mm] {Novel Relationships};

  \node (star1) [star, star points=5, star point ratio=2.25, fill=col3!80, draw, inner sep=0pt, minimum size=8pt, left=of shirt, xshift=-7mm] {};
  \node[star, star points=5, star point ratio=2.25, fill=col3!80, draw, inner sep=0pt, minimum size=8pt, right of=star1, xshift=-5mm] {};
            
        \end{tikzpicture}
        \subcaption{\emph{OvR-SGG}.}
\end{minipage} \vfill 
    \begin{minipage}{\textwidth}
    \vspace{-1mm}
   \centering
    \begin{tikzpicture}[node distance=8mm, auto, scale=\scaleF, every node/.style={scale=\scaleF}]
        \node at (0, 0) {};
        \node (A) at (0.8, 0) {};
        \node (B) at (1.8,-1.) {};
        \node (A1) at (5.8, 0) {};
        \node (C) at (4.8, -1) {};
        
        \draw[line width=3pt, -{Triangle[length=3mm, width=3mm]}, color=color_flow] (A) to (B);
        \draw[line width=3pt, -{Triangle[length=3mm, width=3mm]}, color=color_flow] (A1) to (C);
    \end{tikzpicture}
    \end{minipage}
        \vfill 
\begin{minipage}{\textwidth}
        \centering
        \begin{tikzpicture}[node distance=8mm, auto, scale=\scaleF, every node/.style={scale=\scaleF}]

        \node (img1) [] {\includegraphics[width=0.7in]{pics/2315830.jpg}};
        \node (box1) [right=of img1, xshift=15mm, sggnode] {SGG Model}; 
        \node (cap1) [above =of img1, yshift=-12mm]{\emph{a man \colorbox{pink}{wearing} a shirt is \colorbox{blue!30}{riding} a \colorbox{color_sk}{skateboard}}}; 
        \draw [myarrow] (img1) to (box1);
        \draw [myarrow] (box1) to ([xshift=15mm]box1.east);
        \draw[->, dashed, rounded corners=3pt] (cap1.east) --
        ([xshift=7.5mm]cap1.east) -- (box1.north); 
        \node [right=of cap1, xshift=-11mm, yshift=3mm] {Novel Objects / Relationships}; 
        
            \node[draw, circle, fill=color_man, right=of box1, xshift=30mm] (man) {};
            \node[left of=man, xshift=3mm] {man}; 
            \node[draw, circle, below right =of man, fill=color_pant] (pant) {};
            \node[right of=pant, xshift=-2mm] {pant}; 
            \node[draw, circle, below left=of man, fill=color_helmet] (helmet) {};
            \node[left of=helmet, xshift=0mm] {helmet};
            \node[draw, circle, above =of man, fill=color_shirt] (shirt) {};
            \node[right of=shirt, xshift=-2mm] {shirt}; 
            \node[draw, circle, thick, densely dashed, right =of man, fill=color_sk] (skateboard) {};
            \node[below of=skateboard, xshift=4mm, yshift=5mm] {skateboard}; 
            \node[draw, circle,  above =of skateboard, fill=color_wheel] (wheel) {};
            \node [right of=wheel, xshift=-2mm] {wheel}; 
            
            \draw[->, thick, densely dashed, >=stealth] (man) -- node[midway, sloped, anchor=center, above] {\textcolor{blue!30}{riding}} (skateboard);
            \draw[->, >=stealth] (wheel) -- node[midway, sloped, anchor=center, above] {on} (skateboard);
            \draw[->, >=stealth, thick, densely dashed] (man) -- node[midway, sloped, anchor=center, above] {\textcolor{pink}{wearing}} (shirt);
            \draw[->,>=stealth] (man) -- node[midway, sloped, anchor=center, above] {in} (pant);
            \draw[->, >=stealth] (helmet) -- node[midway, sloped, anchor=center, above] {on} (man);

   \node (star1) [star, star points=5, star point ratio=2.25, fill=col3!80, draw, inner sep=0pt, minimum size=8pt, left of=shirt, xshift=-14mm] {};
  \node (star2) [star, star points=5, star point ratio=2.25, fill=col3!80, draw, inner sep=0pt, minimum size=8pt, right of=star1, xshift=-5mm] {};
  \node[star, star points=5, star point ratio=2.25, fill=col3!80, draw, inner sep=0pt, minimum size=8pt, right of=star2, xshift=-5mm] {};

        \end{tikzpicture}
        \subcaption{\emph{OvD+R-SGG}. }
\end{minipage}
\caption{Illustration of SGG Scenarios (best view in color). Dashed nodes or edges in (a) - (d) refer to unseen category instances (during training), and stars refer to the difficulty of each setting. 
Previous works \cite{xu2017scene, zellers2018neural, tang2019learning, tang2020unbiased, chiou2021recovering, li2021bipartite, chen2019knowledge, zhang2019graphical} mainly focus on \emph{Closed-set SGG} and few studies \cite{he2022towards, zhang2023learning} cover \emph{OvD-SGG}. 
In this work, we give a more comprehensive study towards fully open vocabulary SGG.
}
\label{fig:scenarios}
\end{figure*}
\IEEEPARstart{S}{cene} Graph depicts visual objects and their inter-relationships in a scene.
This structured representation has gained increasing attention and has been utilized in various visual tasks, including image captioning~\cite{yang2019auto, chen2020say, gu2019unpaired, wang2019role, nguyen2021defense}, visual question answering~\cite{teney2017graph, nuthalapati2021lightweight, kenfack2020robotvqa, lee2019visual}, image generation~\cite{johnson2018image, yang2022diffusion}, and robot navigation~\cite{werby2024hierarchical,yin2024sg,miao2024sgloc}, \textit{etc.}
The study of Scene Graph Generation (SGG) has achieved significant advancements, 
including standard benchmark datasets like Visual Genome~\cite{xu2017scene}, 
classical frameworks like IMP~\cite{xu2017scene}, and Motifs~\cite{zellers2018neural}.
Despite such achievements, 
current methods primarily work within a limited framework, restricting object and relationship categories to a predefined set. 
This constraint restricts the wider applicability of SGG models across various real-world applications.

This drawback of prevalent SGG models originates in two aspects. 
First, the complex and costly annotation of scene graphs leads to a shortage of large-scale scene graph datasets with a rich vocabulary.
To alleviate this issue, some weakly-supervised methods~\cite{zhong2021learning,li2022integrating} 
utilize a language scene parser to obtain ungrounded relationship triplets as pseudo labels for learning scene graphs. 
However, as pointed out in \textit{GPT4SGG}~\cite{chen2023gpt4sgg},
these methods suffer from inaccurate parsing results, ambiguous grounding of relationship triplets, and biased caption data.
Second, prevalent SGG models treat object and relationship recognition as a multi-class classification problem, typically relying on a fixed classifier, such as a fully connected layer, in such cases.
Inspired by advancements in open-vocabulary object detection \cite{zareian2021open, wu2023aligning, li2022grounded, zhong2022regionclip, du2022learning}, recent studies \cite{he2022towards, zhang2023learning} have sought to augment the SGG task from a closed-set to an open-vocabulary domain. 
However, these efforts primarily adopt an object-centric perspective, focusing solely on scene graph nodes.
A holistic approach to open-vocabulary SGG requires a thorough analysis of both nodes and edges. 
This raises two key questions that drive our research:
\textit{\textbf{Can the model predict unseen objects or relationships ?}}
\textit{\textbf{What happens when the model encounters both unseen objects and unseen relationships in a scene?}}

To address the drawback, 
we compare different pipelines for relation-aware pre-training, where only weak supervision of scene graphs is provided. 
In this work, we evaluate three types of pipelines:
(1) a \emph{scene parser-based pipeline}~\cite{mao2018parser}, 
which leverages caption data and a natural language parser to extract ungrounded relationship triplets; 
(2) an \emph{LLM (Large Language Model)-based pipeline}, such as \textit{GPT4SGG}~\cite{chen2023gpt4sgg}, which replaces the scene parser with a more powerful LLM to synthesize a dense scene graph; 
and (3) a \emph{multimodal LLM-based pipeline}, which directly generates a dense scene graph with grounded objects. 
These pipelines enable the learning of transferable scene graph knowledge with minimal or no human annotations, 
significantly enhancing SGG model performance, 
particularly in open-vocabulary settings.

Building on our analysis of the limitations in current SGG models and the pipelines for relation-aware pretraining,
we re-evaluate the conventional SGG settings and propose four distinct scenarios (see Fig. \ref{fig:scenarios}):
\begin{enumerate}
    \item \textbf{\textit{Closed-set SGG}}: Having been extensively studied in previous works \cite{xu2017scene,zellers2018neural,tang2019learning,tang2020unbiased,chiou2021recovering,li2021bipartite,chen2019knowledge,zhang2019graphical}, this setting involves predicting nodes (\emph{i.e.}, objects) and edges (\emph{i.e.}, relationships) from a predefined set. Research in this area mainly focuses on feature aggregation and unbiased learning to mitigate long-tail distribution issues.
    \item \textbf{\textit{OvD-SGG}} (Open Vocabulary object Detection-based SGG): 
    As explored in \cite{zhang2023learning}, 
    this scenario extends \textit{Closed-set SGG} from the object (node) perspective by enabling the recognition of unseen object categories during inference. 
    However, the set of relationships remains fixed.
    
    \item \textbf{\textit{OvR-SGG}} (Open Vocabulary Relation-based SGG): 
    This scenario introduces an open-vocabulary setting for relationships, 
    requiring the model to predict unseen relation categories. 
    This task is inherently more challenging due to the absence of pre-trained relation-aware models and reliance on less accurate scene graph annotations. 
    Specifically, while \textit{OvD-SGG} omits unseen object categories during training—yielding graphs with fewer nodes but correct edges—\textit{OvR-SGG} excludes unseen relation categories,
    resulting in graphs with fewer edges. Consequently, the model must distinguish novel relationships from the ``background''.
    
    \item \textbf{\textit{OvD+R-SGG}} (Open Vocabulary Detection+Relation-based SGG):
    The most challenging scenario, where both unseen objects and unseen relationships are present, 
    leads to sparser and less accurate training graphs. 
    This setting demands robust detection and relation modeling under open-vocabulary conditions.
\end{enumerate}
These distinct scenarios exhibit unique intrinsic characteristics and challenges, 
motivating further investigation into open-vocabulary scene graph generation.

With these distinct challenges in mind, 
we introduce \emph{\textit{OvSGTR}} (\emph{Open-vocabulary Scene Graph Transformers}), 
a novel framework designed to tackle the complexities of open-vocabulary SGG.
Our approach not only predicts unseen objects and relationships 
but also effectively handles the scenario 
in which both nodes and edges correspond to categories absent during training. 
Our framework consists of three main components: (1) a frozen image backbone for visual feature extraction, 
(2) a frozen text encoder for textual feature extraction, 
and (3) a transformer decoder for scene graph generation. 
During relation-aware pre-training,  we explored three pipelines to generate scene graph supervision. 
In the fine-tuning phase, relation triplets with location information (\emph{i.e.}, bounding boxes) are sampled from manual annotations. These triplets are then associated with visual features, and visual-concept similarities are computed for both nodes and edges. The resulting similarities are used to predict object and relation categories, thereby enhancing the model's ability to generalize to unseen categories.

Furthermore, in our evaluation of relation-involved open-vocabulary SGG settings (\emph{i.e.}, \textit{OvR-SGG} and \textit{OvD+R-SGG}), 
we empirically identified a catastrophic forgetting problem concerning relation categories. 
This forgetting degrades the model's ability to retain information learned from the relation-aware pre-training stage 
when it is subsequently exposed to fine-grained SGG annotations. 
To address this challenge, we propose a visual-concept retention mechanism coupled with a knowledge distillation strategy. 
In our approach, a pre-trained model serves as a teacher to guide the student model, 
ensuring that the rich semantic space of relations is preserved while adapting to new, fine-grained annotations. 
Simultaneously, the visual-concept retention mechanism helps maintain the model's proficiency in recognizing novel relations.

To our knowledge, \emph{OvSGTR} is the first framework towards fully open-vocabulary SGG. 
We hope that this work will inspire a series of follow-up studies to further advance open-vocabulary scene graph generation. 
In summary, our key contributions are as follows:
\begin{itemize}
\item   We conduct an in-depth study of open-vocabulary SGG from both node and edge perspectives, distinguishing four distinct settings—\textit{Closed-set SGG}, \textit{OvD-SGG}, \textit{OvR-SGG}, and \textit{OvD+R-SGG}. Our quantitative and qualitative analyses provide a holistic understanding of the unique challenges inherent to each scenario.
\item   We introduce a novel SGG framework, \emph{OvSGTR},  where both objects (nodes) and relationships (edges) are extendable to unseen categories, 
greatly broadening the practical applicability of SGG models in real-world scenarios.
\item  By exploring three relation-aware pre-training pipelines, 
our approach significantly enriches the representation of relationships in open-vocabulary settings.
\item Extensive experiments on the VG150 benchmark validate the effectiveness of our framework, demonstrating state-of-the-art performance across all evaluated scenarios.
\end{itemize}
\section{Related Work}
\emph{Scene Graph Generation (SGG)} aims to produce informative graphs that localize objects and describe the relationships among them. 
Prior work has largely focused on contextual information aggregation \cite{xu2017scene,DBLP:conf/iccv/LiOZWW17,zellers2018neural,tang2019learning,DBLP:conf/cvpr/ChenYCL19,li2022sgtr,DBLP:conf/nips/KhandelwalS22,DBLP:journals/pami/CongYR23} and unbiased learning to address the long-tail distribution \cite{tang2020unbiased,chiou2021recovering,li2021bipartite,DBLP:journals/pami/SunZLHL23,DBLP:conf/cvpr/JinGMZXWMS23,DBLP:conf/eccv/JeonKYP24}. 
Typically, SGG methods rely on closed-set object detectors such as Faster R-CNN \cite{ren2015faster}, 
which are unable to handle unseen objects or relations, thereby limiting their real-world applicability. 
Recent studies \cite{he2022towards,zhang2023learning} have extended closed-set SGG to object-centric open vocabulary SGG; 
however, these approaches still struggle to generalize to unseen relations as well as combinations of unseen objects and relations.

An alternative approach to advancing SGG involves leveraging weak supervision from image caption data, 
leading to the emergence of \emph{language-supervised SGG} \cite{zhong2021learning,li2022integrating,zhang2023learning}. 
This strategy offers a cost-effective alternative to expensive manual annotation. 
Nonetheless, prior work using language supervision faces notable limitations: 
(1) it is confined to closed-set relation recognition, and 
(2) issues such as object grounding ambiguity and the inherently biased, 
sparse nature of caption data often yield incomplete scene graphs, as highlighted in \emph{GPT4SGG} \cite{chen2023gpt4sgg}.

In contrast, our framework is fully open vocabulary. 
It overcomes the closed-set assumptions of previous methods \cite{zhong2021learning,zhang2023learning}, 
enabling the model to learn rich semantic concepts that generalize to downstream tasks. 
Moreover, we provide a comprehensive study on relation-aware pretraining, 
an area that has been underexplored in prior SGG research.

In essence, our work generalizes open vocabulary SGG by seamlessly integrating it with closed-set SGG. 
To our knowledge, this represents the first unified framework dedicated to achieving fully open vocabulary SGG, 
encompassing both the nodes and edges of scene graphs.
\begin{figure*}[t]
    \centering
    \includegraphics[width=0.985\textwidth]{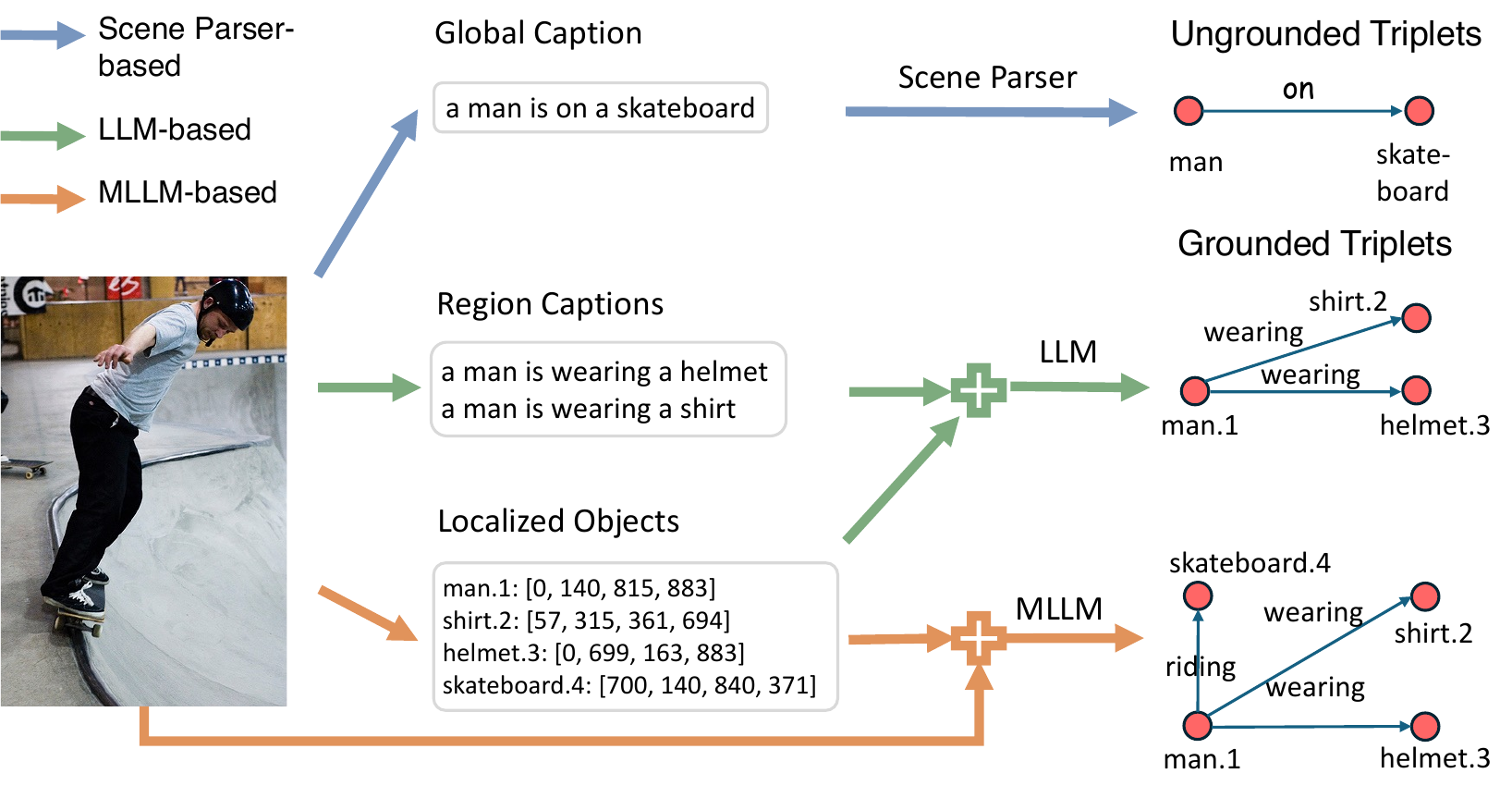}
    \caption{Comparison of different pipelines for relation-aware pre-training. (a) Early weakly-supervised methods~\cite{zhong2021learning,li2022integrating} utilize a language scene parser~\cite{mao2018parser} to extract relationship triplets; 
    (b) LLM-based methods replace the scene parser with a more powerful LLM to synthesize a more dense scene graph, \eg, \emph{GPT4SGG}~\cite{chen2023gpt4sgg};
    (c) Multimodal LLM (MLLM)-based pipeline directly digests an input image and outputs a dense scene graph, \eg, MegaSG~\cite{chen2024makes}. 
    }
    \label{fig:pipelines}
\end{figure*}

\emph{Vision-Language Pretraining (VLP)} has recently attracted considerable attention in a wide range of vision-language tasks. 
The primary objective of VLP is to establish an alignment between visual and linguistic semantic spaces. 
For example, CLIP \cite{radford2021learning} demonstrates impressive zero-shot image classification capabilities through contrastive learning on large-scale image-text datasets. 
Building upon this success, several subsequent methods \cite{li2022grounded, liu2023grounding, zhong2022regionclip} have focused on learning fine-grained alignments between image regions and language data, 
thereby enabling object detectors to recognize unseen objects by leveraging linguistic information. 
This success on downstream tasks exemplifies the potential of aligning visual features with relational concepts, 
which is a fundamental step towards developing a fully open-vocabulary scene graph generation framework.

\emph{Open-vocabulary Object Detection (OvD)} aims to detect previously unseen classes during inference, 
thereby breaking the constraint of a fixed, predefined object set (\eg, 80 categories in COCO~\cite{chen2015microsoft}). 
To achieve this, Ov-RCNN \cite{zareian2021open} transfers semantic knowledge acquired from image captions to the downstream object detection task. 
Notably, supervision signals for unseen or novel classes are excluded during detector training, 
even though these classes are present in the large vocabulary of captions. 
Beyond direct caption learning, ViLD \cite{DBLP:conf/iclr/GuLKC22} distills knowledge from a pre-trained open-vocabulary image classification model into a two-stage object detector. 
In addition to OvD, various methods and applications have been developed, 
including open-vocabulary segmentation \cite{DBLP:conf/eccv/GhiasiGCL22}, open-vocabulary video understanding \cite{DBLP:journals/corr/abs-2109-08472}, 
and open-vocabulary scene graph generation \cite{he2022towards, zhang2023learning, li2024pixels}. 
For a more comprehensive analysis of open-vocabulary learning, 
we refer readers to \cite{wu2024towards} and \cite{DBLP:journals/corr/abs-2307-09220}.

\emph{Large Language Models (LLMs)} have been emerged as a powerful technique that widely affects the research community. 
Its complex reasoning and zero-shot capabilities provides a new paradigm for visual reasoning tasks like \emph{SGG}. 
For instance, LLM4SGG~\cite{kim2024llm4sgg} employed LLMs to extract relational triplets from captions through original and paraphrased text.
GPT4SGG~\cite{chen2023gpt4sgg} leveraged an LLM to synthesize scene graphs from local and global textual descriptions. 
Beyond LLMs, the multimodal LLMs like GPT-4V~\cite{OpenAI2023GPT4V} show strong visual reasoning capabilities. 
These drive us to build a connection between the relation-aware pre-training and (multimodal) LLMs, in which how to leverage the (multimodal) LLM become the critical question for learning a transferable visual knowledge. 
\section{Methodology}
\input{pics/framework}
Given an image $I$, the goal of the SGG task is to generate a descriptive graph $\mathcal{G} = (\mathcal{V}, \mathcal{E})$, 
where each node $v_i \in \mathcal{V}$ contains location (bounding box) and object category information, and each edge $e_{ij} \in \mathcal{E}$ represents the relationship between nodes $v_i$ and $v_j$. 
In open-vocabulary settings, the label set $\mathcal{C}$ is divided into two disjoint subsets: 
\emph{base classes} $\mathcal{C}_B$ and \emph{novel classes} $\mathcal{C}_N$ $(\mathcal{C}_B \cup \mathcal{C}_N = \mathcal{C}$, $\mathcal{C}_B \cap \mathcal{C}_N = \emptyset$).

In this work, we unify node- and edge-level open-vocabulary scene graph generation within a single end-to-end transformer framework. 
Compared to our previous conference paper~\cite{chen2024expanding}, we studied three relation-aware pre-training pipelines—namely, a scene parser-based, an LLM-based, and a multimodal LLM-based pipeline—to enrich visual relationship understanding. 
In addition, we expand our experimental evaluation by including comprehensive experiments on the GQA dataset, 
which demonstrate improved generalization in diverse, real-world scenarios.
\subsection{Relation-aware Pre-training}
\label{sec:pretraining}
The common paradigm in SGG is to use a pre-trained object detector (\emph{e.g.}, Faster R-CNN) to provide a good initial for detecting objects, while this neglects the relationship between objects. 
To learn a transferable knowledge of visual relationships, 
three pipelines have been explored in this work, as shown in Fig. ~\ref{fig:pipelines}: 
\begin{itemize}
    \item[{(a)}] \textit{Scene Parser-based Pipeline}: {Learning from the caption} as in previous methods~\cite{zhong2021learning,li2022integrating}. 
    These approaches leverage a language parser to extract ungrounded triplets from captions. 
    However, they face challenges such as inaccurate parsing, ambiguity in grounding triplets, 
    and biased as well as sparse caption data (see \emph{GPT4SGG}~\cite{chen2023gpt4sgg} for further discussion).
    \item[{(b)}] \textit{LLM-based Pipeline}: {Learning from the LLM} as in \textit{GPT4SGG}~\cite{chen2023gpt4sgg}. This method adopts a two-stage pipeline: first, captioning dense regions, and then synthesizing the scene graph using a large language model (\emph{e.g.}, GPT-4). The principal advantage of this pipeline is its powerful language understanding and reasoning capabilities, intrinsic to LLMs, 
    which enable it to overcome limitations (\emph{e.g.}, inaccurate parsing) of conventional scene parser-based pipeline.
    \item[{(c)}] \textit{Multimodal LLM-based Pipeline}: {Learning from the multimodal LLM} as in MegaSG~\cite{chen2024makes}. 
    This approach directly processes an input image to generate a dense scene graph. 
    A potential drawback is the limited spatial information (\emph{i.e.}, bounding boxes) provided, 
    as multimodal LLMs are less effective at localizing objects compared to dedicated object detectors.
    Nonetheless, unlike an LLM-based pipeline, this paradigm directly captures visual details, thereby enabling a more precise association between nodes and edges.   
\end{itemize}

These pipelines aim to learn knowledge about visual relationships without or with fewer human annotations,
significantly boosting the performance of SGG models under different settings.
\subsection{Fully Open Vocabulary Architecture}
As shown in Fig.~\ref{fig:framework}, 
\emph{OvSGTR} is a DETR-like architecture composed of three main components: 
a visual encoder for image feature extraction, 
a text encoder for processing textual data, 
and a transformer that simultaneously performs object detection and relationship recognition. 
When provided with paired image-text data, \emph{OvSGTR} adeptly generates the corresponding scene graphs. 
To reduce the optimization burden, the weights of both the image backbone and the text encoder are frozen during training.

\emph{Feature Extraction.} Given an image--text pair, 
we extract multi-scale visual features using an image backbone (\emph{e.g.}, Swin Transformer~\cite{liu2021swin}) and obtain text features via a text encoder (\emph{e.g.}, BERT~\cite{kenton2019bert}). 
The visual and text features are then fused and enhanced using cross-attention within the deformable encoder module of the transformer.

\emph{Prompt Construction.} We construct the text prompt by concatenating all possible (or sampled) noun phrases and relation categories. 
For example, a prompt may be formatted as \texttt{[CLS] girl. umbrella. table. bathing. $\cdots$ zebra. [SEP]  on. in. wears. $\cdots$ walking. [SEP][PAD][PAD]},
which is analogous to the strategy employed in GLIP~\cite{li2022grounded} and Grounding DINO~\cite{liu2023grounding} that concatenate all noun phrases. For a large vocabulary during training, we randomly sample negative words and constrain the total number of positive and negative words to be $M$ (\emph{e.g.}, $M=80$).

\emph{Node Representation.} Given $K$ object queries, the model follows the standard DETR framework to output $K$ hidden features $\{\bm{v}_i\}_{i=1}^{K}$. Each feature is processed by:
\begin{itemize}
    \item a bounding box head (a three-layer fully connected network) to decode location information (\emph{i.e.}, a 4-dimensional vector), and
    \item a parameter-free classification head that computes the similarity between hidden features and text features.
\end{itemize}
These hidden features serve as the visual representations for the predicted nodes.

\emph{Edge Representation.} Instead of using a complex message passing mechanism for relation features, 
we design a lightweight relation head that concatenates the node features for the subject and object with relation query features. 
To learn a relation-aware representation, we initialize a random embedding for querying relations. 
This embedding interacts with the image and text features via cross-attention in the decoder stage. 
Based on this design, for any subject--object pair $(s_i, o_j)$, 
the edge representation is computed as
\begin{equation}
    \bm{e}_{s_i \rightarrow o_j} = f_{\theta}\Big([\bm{v}_{s_i},\, \bm{v}_{o_j},\, \bm{r}]\Big),
    \label{eq:rln}
\end{equation}
where $\bm{v}_{s_i}$ and $\bm{v}_{o_j}$ denote the node representations for the subject and object, respectively, 
$\bm{r}$ represents the relation query features, $[\cdot]$ indicates concatenation, and $f_{\theta}$ is a two-layer multilayer perceptron.

\emph{Loss Function.} 
Following previous DETR-like methods~\cite{zhu2021deformable, liu2023grounding}, 
we employ an L1 loss and a GIoU loss~\cite{rezatofighi2019generalized} for the bounding-box regression. 
For object or relation classification, 
we use Focal Loss~\cite{lin2017focal} as a contrastive loss between predictions and language tokens.

To decode object and relation categories in a fully open-vocabulary manner, the fixed classifier (a fully connected layer) is replaced with a visual--concept alignment module (detailed in Section~\ref{sec:lvca}).
\subsection{Learning Visual-Concept Alignment}
\label{sec:lvca}
A central component of our framework is the alignment of visual features with their corresponding textual concepts. 
At the node level, given an image, our model produces $K$ predicted nodes $\{\tilde{v}_j\}_{j=1}^{K}$, 
which must be accurately paired with the $N$ ground-truth nodes $\{v_i\}_{i=1}^{N}$. 
To achieve this, we employ a bipartite matching strategy akin to that used in standard DETR. 
The matching process is formulated as
\begin{equation}
    \max_{\mathbf{M}} \sum_{i=1}^{N}\sum_{j=1}^{K} \mathrm{sim}(v_i,\tilde{v}_j) \cdot \mathbf{M}_{ij},
    \label{eq:bipart}
\end{equation}
where $\mathrm{sim}(\cdot,\cdot)$ measures the similarity between a ground-truth node and a predicted node, 
taking into account both spatial (\emph{e.g.}, bounding box) and semantic (\emph{e.g.}, category) cues. 
Here, $\mathbf{M} \in \mathbb{R}^{N\times K}$ is a binary assignment matrix with $\mathbf{M}_{ij}=1$ 
if node $v_i$ is matched with $\tilde{v}_j$, and $\mathbf{M}_{ij}=0$ otherwise.

For each matched pair $(v_i, \tilde{v}_j)$, we enhance their alignment by jointly optimizing spatial and semantic consistencies.
The spatial alignment is enforced by minimizing the L1 and GIoU losses between the corresponding bounding boxes. 
Concurrently, the categorical similarity is defined as
\begin{equation}
    \mathrm{sim}_{\mathrm{cat}}(v_i,\tilde{v}_j) = \sigma\big(\langle \mathbf{w}_{v_i}, \mathbf{v}_j \rangle\big),
    \label{eq:align1}
\end{equation}
where $\mathbf{w}_{v_i}$ is the word embedding corresponding to the ground-truth node $v_i$, 
$\mathbf{v}_j$ is the visual representation of the predicted node $\tilde{v}_j$, $\langle \cdot,\cdot \rangle$ denotes the dot product, and $\sigma(\cdot)$ is the sigmoid function. 
This formulation serves to align the visual features of nodes with their semantic prototypes in the text space.

To extend relation recognition from a closed-set to an open-vocabulary setting, 
we further learn a joint visual-semantic space that aligns relation features with textual descriptions. 
Specifically, given a text input $\mathbf{t}$ processed by a text encoder $E_t$, 
and a relation feature $\mathbf{e}$, we compute the alignment score as
\begin{equation}
    s(\mathbf{e}) = \langle \mathbf{e},\, f(E_t(\mathbf{t})) \rangle,
\end{equation}
where $f(\cdot)$ is a fully connected layer that projects the text embedding into the relation feature space. 
The alignment is supervised via a binary cross-entropy loss defined as
\begin{equation}
\begin{split}
        \mathcal{L}_{\mathrm{bce}} =& \frac{1}{|\mathcal{P}|+|\mathcal{N}|} \sum_{\mathbf{e} \in \mathcal{P}\cup\mathcal{N}} \Bigl[- y_{\mathbf{e}} \log \sigma\big(s(\mathbf{e})\big) - \\
        &\bigl(1-y_{\mathbf{e}}\bigr) \log \Bigl(1-\sigma\big(s(\mathbf{e})\Bigr)\Bigr],
\end{split}
    \label{eq:bce}
\end{equation} 
where $y_{\mathbf{e}} \in \{0,1\}$ indicates whether $\mathbf{e}$ corresponds to a positive sample (with positive and negative samples denoted by $\mathcal{P}$ and $\mathcal{N}$, respectively) and $\sigma(\cdot)$ is the sigmoid function. 
This loss encourages high alignment scores for true correspondences, while penalizing false matches.

Learning visual-concept alignment is inherently challenging due to the scarcity of relation-aware pre-trained models on large-scale datasets. 
While object-language alignment has advanced considerably with models like CLIP \cite{radford2021learning} and GLIP \cite{li2022grounded}, relation modeling remains largely underexplored. 
Moreover, the manual annotation of scene graphs is both labor-intensive and costly, 
limiting the scalability of SGG datasets. 
To address these challenges, we investigate three paradigms of relation-aware pre-training (see Section \ref{sec:pretraining}). 
By leveraging weak supervision, our approach facilitates the learning of rich representations for both objects and relationships.
\subsection{Visual-Concept Retention with Knowledge Distillation}
\label{sec:vcp}
To enable the model to recognize a diverse set of objects and relationships beyond a limited fixed vocabulary, 
we build upon the visual-concept alignment introduced in Section~\ref{sec:lvca}. 
Empirically, we observe that directly optimizing the model on a new dataset using Eq.~\ref{eq:bce} leads to catastrophic forgetting—even when starting from a relation-aware pre-trained model. 
This issue is particularly evident in the \textit{OvR-SGG} or \textit{OvD+R-SGG} settings, 
where unseen (or novel) relationships are excluded from the scene graph. 
As a result, the model must discriminate novel relations from the background, a task that becomes increasingly challenging.

To alleviate this problem, we incorporate a knowledge distillation strategy that preserves the learned semantic space. Specifically, we designate the pre-trained model (obtained via weak supervision as described in Section~\ref{sec:pretraining}) as the teacher, while the current model acts as the student. For each negative sample, we enforce that the student’s edge features approximate those of the teacher. Formally, the distillation loss is defined as
\begin{equation}
    \mathcal{L}_{\mathrm{distill}} = \frac{1}{|\mathcal{N}|} \sum_{\bm{e} \in \mathcal{N}} \left\| \bm{e}^s - \bm{e}^t \right\|_1,
    \label{eq:distill}
\end{equation}
where $\bm{e}^s$ and $\bm{e}^t$ denote the student’s and teacher’s edge features, respectively, and $\mathcal{N}$ represents the set of negative samples. The overall training objective then becomes
\begin{equation}
    \mathcal{L} = \mathcal{L}_{\mathrm{bce}} + \lambda \mathcal{L}_{\mathrm{distill}},
\end{equation}
with $\lambda$ balancing the contributions of the ground-truth supervision and the distillation loss.

This knowledge distillation approach enables the model to acquire more precise information for base classes 
while preserving its capacity to generalize to unseen categories during fine-tuning. 
Consequently, our pipeline—pre-training on large-scale, coarse data followed by fine-tuning on a more fine-grained SGG dataset—
becomes more effective and adaptable.
\section{Experiments}   
\subsection{Datasets and Experiment setup}
\textit{Datasets.}
The VG150 dataset~\cite{xu2017scene} is a widely adopted benchmark that provides human-annotated labels for $150$ object categories and $50$ relation categories. 
It comprises a total of $108,777$ images, with $70\%$ reserved for training, $5,000$ for validation, and the remainder for testing. 
Consistent with the evaluation of $\text{VS}^3$~\cite{zhang2023learning}, we remove images utilized by the pre-trained object detector Grounding DINO~\cite{liu2023grounding}, resulting in a test set of $14,700$ images.

For relation-aware pre-training, we explored three different pipelines:
\begin{itemize}
    \item \textit{Scene Parser-based Pipeline:} We employ an off-the-shelf language parser~\cite{mao2018parser} to extract relation triplets from image captions, yielding ${\sim}104$k images with coarse scene graphs from the COCO caption training set.
    \item \textit{LLM-based Pipeline:} ${\sim}113$k COCO images are annotated with scene graphs by GPT-4~\cite{achiam2023gpt} in \emph{GPT4SGG}~\cite{chen2023gpt4sgg}.
    \item \textit{Multimodal LLM-based Pipeline:} $1$M images are annotated using Gemini 1.5 Flash~\cite{reid2024gemini} in \emph{MegaSG}~\cite{chen2024makes}. To prevent information leakage and reduce the computational burden, we select $644$k images from \emph{MegaSG} for pre-training.
\end{itemize}

GQA~\cite{hudson2019gqa}  utilizes the same image corpus as Visual Genome~\cite{krishna2017visual} but features more diverse annotations, 
comprising $1,703$ object classes and $310$ predicates. 
Similar to VQ150, GQA200~\cite{dong2022stacked,sudhakaran2023vision} is a reduced version of GQA that contains $200$ object classes and $100$ predicate classes.

\textit{Metrics.} We primarily adopt the \emph{SGDet} protocol~\cite{xu2017scene,tang2020unbiased} (also known as \emph{SGGen}) for fair comparisons. 
SGDet generates a scene graph directly from an input image without any predefined object boxes. 
Performance is evaluated using Recall@K (K = 20/50/100), which measures the fraction of correctly predicted relation triplets among the top-K predictions. A triplet is considered correct if its subject, object, and predicate labels match the ground truth, and both subject and object regions have the same label or an IoU $\geq 0.5$. 
To ensure comprehensive evaluation, we also report results for the \emph{PredCls} setting, 
where ground-truth object labels and locations are provided. 
Since our approach detects objects in a one-stage manner, we implement \emph{PredCls} by selecting image regions that best match the ground-truth objects in post-processing before performing relation recognition.
For the \emph{Closed-set SGG} setting, we additionally report Mean Recall@K (mR@K, K = 20/50/100) and inference speed.

\textit{Implementation Details.} Our model is initialized with pre-trained Grounding DINO~\cite{liu2023grounding}. 
The visual backbone (\emph{i.e.}, Swin-T or Swin-B) and the text encoder (\emph{i.e.}, BERT-base~\cite{kenton2019bert}) remain frozen, while other components, including the relation-aware embedding, are randomly initialized. 
For pairwise relation recognition, we retain the top 100 detected objects per image. 
The model is trained on an 8$\times$NVIDIA A100 GPU cluster using the AdamW~\cite{adamw} optimizer. 
\begin{table*}[t]
    \centering
\caption{Zero-shot performance of state-of-the-art methods on the VG150 test set. For the COCO Caption dataset, a language parser \cite{mao2018parser} has been used for extracting triplets from the caption. 
To prevent information leakage, we sampled 644k images from MegaSG, ensuring that the CLIP similarity of each sampled image with the VG test set remained below 0.9.
}
   \resizebox{0.985\textwidth}{!}
    {
    \begin{tabular}{l|c|ccc|ccc|ccc|ccc}
    \toprule
         \multirow{2}{*}{SGG model}&  \multirow{2}{*}{Training Data} &  
         \multicolumn{6}{c|}{SGDet} 
         & \multicolumn{6}{c}{PredCls} 
         \\  
         &  &  
         \multicolumn{3}{c|}{R@20/50/100} & 
         \multicolumn{3}{c|}{mR@20/50/100} &
         \multicolumn{3}{c|}{R@20/50/100} & 
         \multicolumn{3}{c}{mR@20/50/100} 
         \\
    \midrule
         LSWS  \cite{yelinguistic}&   & - & 3.28 & 3.69   
         &    \multicolumn{3}{c|}{-} &
          \multicolumn{3}{c|}{-} &  \multicolumn{3}{c}{-}  
         \\ 
         MOTIFS \cite{zellers2018neural}&       & 5.02 & 6.40 & 7.33   &  \multicolumn{3}{c|}{-}&
          \multicolumn{3}{c|}{-} &  \multicolumn{3}{c}{-}  
         \\ 
         Uniter \cite{chen2020uniter} &COCO \cite{chen2015microsoft}
              & 5.42 & 6.74 & 7.62  &   \multicolumn{3}{c|}{-}&\multicolumn{3}{c|}{-} &  \multicolumn{3}{c}{-}  \\ 
         $\text{VS}^3_{\text{(Swin-T)}}$  \cite{zhang2023learning}   &  Caption   & 4.56 & 5.79 & 6.79  & 2.18 & 2.59 & 3.00 & 12.30 & 16.77
         & 19.40 & 3.56 & 4.79 & 5.51 
         \\ 
         $\text{VS}^3_{\text{(Swin-L)}}$  \cite{zhang2023learning}  &  (104k)     & 4.82 & 6.20 & 7.48  & 2.29  &  2.70  &  3.09
         & 12.54 & 17.28 & 19.89 
         & 3.57 & 4.83 & 5.56 
         \\ 
         $\text{OvSGTR}_\text{(Swin-T)}$   &    &  {{6.61}} & {{8.92}} & {{10.90}}  & 1.09   &  1.53  & 1.95 
         & 16.65 & 22.44 & 26.64 & 2.47 & 3.58 & 4.41 
         \\ 
        $\text{OvSGTR}_\text{(Swin-B)}$   &    &   {6.85}
          &    {9.33}  &  {11.47}   
          &   {1.28}   &   {1.79}
          &  {2.18}  &
          16.82 & 22.79 & 27.04 & 2.94 & 4.24 & 5.26  
         \\ 
      \hline 
       $\text{VS}^3_{\text{(Swin-T)}}$  \cite{zhang2023learning}&    & 4.92  & 6.32
            &7.22&      1.90  &   2.44  &  2.80  &   13.90  & 17.06 & 18.60 &
            3.87 & 4.81 & 5.30 
            \\ 
      $\text{VS}^3_{\text{(Swin-L)}}$  \cite{zhang2023learning}    & COCO & 
          {5.07} &    {7.40}   &    {9.50} &    {1.30} &  {1.93} &   {2.42}
          & 14.53 & 18.72 & 20.73 & 3.87 & 4.91 &5.46 
          \\ 
       $\text{OvSGTR}_\text{(Swin-T)}$   &   GPT4SGG \cite{chen2023gpt4sgg}               &    7.35 & 9.66     &    11.14  &  2.73  & 3.67    & 4.52          & 21.03 &  24.43 &  25.98 & 6.61  & 7.82  &  8.54         \\
       $\text{OvSGTR}_\text{(Swin-B)}$    &  (113K)    
          &  ${ {7.65}}$  & ${ {10.10}}$    &  ${{11.73}}$    &   ${ {2.92}}$ &   ${{3.84}}$  & ${{4.69}}$
          & 21.13 & 24.78 & 26.25 & 7.10 & 8.49 & 9.37 
          \\
       \hline 
        $\text{VS}^3_{\text{(Swin-T)}}$  \cite{zhang2023learning}    &  & 5.56
             &  8.19      &   10.17 &  1.15  &   1.71  & 2.20  & 
             23.81 & 29.64 & 32.18 
             & 4.70 & 5.96 & 6.57 
             \\ 
        $\text{VS}^3_{\text{(Swin-L)}}$  \cite{zhang2023learning}    & {MegaSG} & 9.74
             &  14.80      & 18.80   &  1.57  &  2.71  &3.75 
             & 31.88 & 38.77 & 41.76 & 5.32 & 6.88 & 7.58
             \\ 
        $\text{OvSGTR}_\text{(Swin-T)}$    &   (644k)      
           &  \textbf{9.94}  & \textbf{13.92}    &  \textbf{17.17}    &   \textbf{3.05} &   \textbf{4.03}  & \textbf{4.76} 
           & \textbf{37.12} & \textbf{44.10} & \textbf{47.09} &
           \textbf{8.49} & \textbf{10.22} & \textbf{11.07} \\
        $\text{OvSGTR}_\text{(Swin-B)}$    &     
           &  \textbf{10.36}  & \textbf{14.61}    &  \textbf{18.13}    &   \textbf{3.01} &   \textbf{4.10}  & \textbf{4.98}
           & \textbf{39.04} & \textbf{45.86} & \textbf{48.54} 
           & \textbf{8.45} & \textbf{10.30} & \textbf{11.12} 
           \\   
   \bottomrule
    \end{tabular}
    }
    \label{tab:mega_zeroshot}
\end{table*}  
\begin{table*}[htbp]
    \centering
    \caption{Comparison with state-of-the-art methods on the VG150 test set (under SGDet protocol). 
    The symbol \ding{71} denotes fully supervised methods. 
    ``VG Caption'' is processed using the Scene Parser-based Pipeline, 
    ``VG@GPT4SGG'' follows the LLM-based Pipeline, 
    and ``VG@Gemini'' employs the Multimodal LLM-based Pipeline.
    }    
   \resizebox{0.985\textwidth}{!}
    {
    \begin{tabular}{l|c|c|ccc|ccc}
    \toprule
         SGG model&  Training Data & Grounding & 
         \multicolumn{3}{c|}{R@20/50/100} & 
         \multicolumn{3}{c}{mR@20/50/100} \\
    \midrule
        IMP \cite{xu2017scene} \ding{71}&   &  &  17.73  & 25.48  & 30.71 & 2.66 & 4.10 & 5.29  \\  
        MOTIFS \cite{zellers2018neural} \ding{71} & VG150  &  & 25.40 & 32.34 & 36.80 & 5.80 & 7.38 & 9.04 \\    
        $\text{VS}^3_{\text{(Swin-L)}}$  \cite{zhang2023learning} \ding{71}  & (${\sim}56k$ )   &{-}  &
        27.34 & 36.04 & 40.88 &  4.43 & 6.45 & 7.81 \\      
        $\text{OvSGTR}_\text{(Swin-B)}$   \ding{71} &  & &
         27.75 & 36.44 &  42.35 & 5.24 & 7.41 & 8.98 \\ 
        \midrule    
        IMP \cite{xu2017scene} &    &GLIP-L \cite{li2022grounded}&    6.51 &  9.54  &11.86  &  0.88 &  1.41 &1.89   \\  
         LSWS  \cite{yelinguistic} &   & - & - & 3.85 & 4.04 & - & - & - \\ 
        MOTIFS \cite{zellers2018neural} & VG Caption &
        Li \emph{et al.}  \cite{li2022integrating} 
        & 8.25 & 10.50 & 11.98 & - & - & - \\ 
        MOTIFS \cite{zellers2018neural} & (${\sim}73k$)  &
        GLIP-L \cite{li2022grounded}  & 
        13.88 & 18.46  & 21.81  &   3.71 & 4.85  & 5.58    \\ 
        $\text{VS}^3_{\text{(Swin-L)}}$  \cite{zhang2023learning}
        &  & GLIP-L \cite{li2022grounded}
        &11.31 &16.00 &19.85& 2.39 & 3.80 & 4.87\\ 
        $\text{OvSGTR}_\text{(Swin-B)}$   &  
         &  GLIP-L \cite{li2022grounded}   &  16.36 & 22.14 & 26.20 & 3.80 &5.24 & 6.25   \\   
        \hline 
        IMP \cite{xu2017scene} &   &  &  12.78  & 16.63  & 19.39  & 3.86  & 4.88  & 5.76  \\ 
        MOTIFS \cite{zellers2018neural} & VG@GPT4SGG  &
&16.41  & 20.46  & 23.23  & 5.86  & 7.36  & 8.51 \\ 
       $\text{VS}^3_{\text{(Swin-L)}}$  \cite{zhang2023learning}
        & (${\sim}46k$) &   
&17.77  & 22.42  & 25.29  & 4.24  & 5.82  & 6.97  \\
         $\text{OvSGTR}_\text{(Swin-B)}$   &   & \multirow{-4}{*}{-}
& 20.12  & 25.03  & 28.84  & 5.68  & 7.14  & 8.22 \\ 
         \hline
         %
       $\text{VS}^3_{\text{(Swin-L)}}$  \cite{zhang2023learning}
        &  VG@Gemini &  & 17.10 & 22.70 & 26.10 & 3.60 & 5.11 & 6.26 
         \\       
         $\text{OvSGTR}_\text{(Swin-B)}$   &  (${\sim}50k$) & \multirow{-2}{*}{-}
         &   19.78 &  26.08  & 30.66  &  5.07 &  6.72 &  7.86  \\        
   \bottomrule
    \end{tabular}
    }
    \label{tab:pretrain_vg}
\end{table*} 
\subsection{Main Results}
\subsubsection{Relation-aware Pre-training}
We systematically evaluate the zero-shot performance of recent state-of-the-art SGG methods under three different large-scale pre-training pipelines (see Section~\ref{sec:pretraining}).
As summarized in Table~\ref{tab:mega_zeroshot}, large-scale pre-training yields consistent improvements in visual relationship recognition across a variety of models.
In particular, our proposed \emph{OvSGTR} (Swin-B) achieves substantial performance gains: \eg, 
it improves R@20, R@50, and R@100 in the PredCls setting from baseline values of 16.82, 22.79, and 27.04 to 39.04, 45.86, and 48.54, respectively. 
These improvements demonstrate the effectiveness of incorporating multimodal knowledge for relation-aware representation learning.

The superior results of \emph{OvSGTR} are largely attributed to two key factors. First, the multimodal LLM-based pipeline produces high-quality scene graph annotations, leading to richer supervision signals for pre-training. 
Second, the use of large-scale data  exposes the model to a diverse range of object and predicate instances, 
improving its ability to generalize to previously unseen visual relationships. 
The zero-shot improvements suggest that the learned representations capture more fine-grained semantic cues, enabling accurate predicate classification even when the specific  \texttt{object-predicate-object} combinations were not explicitly encountered during training.

Moreover, the comparison in Table~\ref{tab:pretrain_vg} shows that these relation-aware pre-training pipelines consistently improve SGG performance on the standard VG150 benchmark. 
The results demonstrate that large-scale, relation-centric pre-training strategies can effectively address the inherent data scarcity in SGG tasks. 
Compared with other baselines like ~\cite{zellers2018neural,tang2020unbiased}, our \emph{OvSGTR} exhibits clear advantages in capturing visual relationships between objects, 
highlighting the importance of leveraging richer linguistic and visual cues during training. 
Overall, these findings underscore the idea that combining large-scale multimodal pre-training with specialized SGG methods can significantly enhance the recognition of complex visual relationships. 
By integrating knowledge from pre-trained language models and carefully curated scene graph annotations, \emph{OvSGTR} achieves state-of-the-art performance, 
enhancing the recognition of complex visual relationships.

\begin{table*}[t]
    \centering
    \caption{Experimental results of \textit{Closed-set SGG} on VG150 test set. ``40M/177M'' in Params. refers to 40M trainable parameters and 177M total parameters.
    Inference time is benchmarked on an NVIDIA RTX 3090 GPU with batch size 1 and an input resolution $1000\times 600$.
    Time for \text{SGNLS} \cite{zhong2021learning} is benchmarked on an NVIDIA A100 GPU (80G) due to memory out of usage.
    Throughout the paper, we use $\star$ to indicate relation-aware pre-training on MegaSG.
    }
    \resizebox{0.985\textwidth}{!}{
    \begin{tabular}{l|c|c|c|>{\centering\arraybackslash}p{8mm}>{\centering\arraybackslash}p{8mm}>{\centering\arraybackslash}p{8mm}
    |>{\centering\arraybackslash}p{7mm}>{\centering\arraybackslash}p{7mm}>{\centering\arraybackslash}p{7mm}|c}
   \toprule
         SGG model& Backbone & Detector& Params. & 
         \multicolumn{3}{c|}{R@20/50/100} & 
         \multicolumn{3}{c|}{mR@20/50/100} &  Time (s)   \\ 
        \midrule 
          IMP \cite{xu2017scene} &  RX-101  & & 146M/308M &  17.7 & 25.5  & 30.7 &2.7  &4.1 &5.3  &  0.25\\
         MOTIFS \cite{zellers2018neural} & RX-101&Faster &205M/367M & 25.5 & 32.8 & 37.2&   5.0& 6.8& 7.9 &   0.27\\
         VCTREE \cite{tang2019learning} &   RX-101& R-CNN &197M/358M & 24.7 & 31.5 & 36.2&  - & - & - & 0.38\\
         \text{SGNLS} \cite{zhong2021learning} &   RX-101&  &  165M/327M  & 24.6 & 31.8 & 36.3&  -& -& -  & $>$ 7\\
         HL-Net \cite{lin2022hl} &    RX-101& & 220M/382M& 26.0 & 33.7 & 38.1& -  &- &-  & 0.10 \\
        \hline
         FCSGG \cite{liu2021fully} &   HRNetW48& - & 87M/87M  &13.6 & 18.6 & 22.5&  2.3 & 3.2& 3.9 &  0.13\\
         SGTR \cite{li2022sgtr} &   R-101& DETR &  36M/96M & - &24.6 & 28.4& - & -&  -&   0.21\\
         $\text{VS}^{3}$ \cite{zhang2023learning} &   Swin-T& -& 93M/233M&26.1 & 34.5 & 39.2& -
          & -& -  & 0.16\\
         $\text{VS}^{3}$ \cite{zhang2023learning} &   Swin-L& -  & 124M/432M & 27.3 & 36.0 & 40.9&  4.4 &6.5 &7.8  & 0.24 \\       
         \hline 
         
         \text{OvSGTR} &   Swin-T& \multirow{4}{*}{DETR}  & 
          41M/178M
         &{27.0} & {35.8} & {41.3}& 
         {5.0} &{7.2} & {8.8} & 0.13\\
         
         \text{OvSGTR} &   Swin-B&   & 
         41M/238M
         & {27.8} & {36.4} & {42.4} & {5.2} & {7.4}&  {9.0} &  0.19 \\
         
         $\text{OvSGTR}^{\star}$ &   Swin-T&  & 
          41M/178M
         &   27.3 &  36.2 & 41.9 & 5.3
          &7.7  & 9.4  & 0.13\\
          
         $\text{OvSGTR}^{\star}$ &   Swin-B&   & 
         41M/238M
         & \textbf{28.6} & \textbf{37.6} & \textbf{43.4} & \textbf{5.8} & \textbf{8.3}&  \textbf{10.2} &  0.19 \\ 
         \bottomrule 
    \end{tabular} }
    \label{tab:closed}
\end{table*}
\begin{table*}[t]
    \centering
    \caption{Experimental results (R@20/50/100) of \textit{OvD-SGG} setting on VG150 test set.
    }    
    \resizebox{0.985\textwidth}{!}{
    \begin{tabular}{l|>{\centering\arraybackslash}p{24mm}>{\centering\arraybackslash}p{24mm}|>{\centering\arraybackslash}p{24mm}>{\centering\arraybackslash}p{24mm}
    } 
            \toprule
            \multirow{2}{*}{Method} & \multicolumn{2}{c|}{Base+Novel (Object)} & 
           \multicolumn{2}{c}{Novel (Object)}  \\  
              & PredCls &  SGDet &   PredCls & SGDet 
              \\ 
          \midrule 
            IMP \cite{xu2017scene} &  46.79 / 54.21 / 56.85 & 2.09 / 2.85 / 3.45 & 42.02 / 51.15 / 54.36 & 0.00 / 0.00 / 0.00 \\ 
            MOTIFS \cite{zellers2018neural} & 45.88 / 53.54 / 56.23 & 2.61 / 3.30 / 3.83 & 39.58 / 49.63 / 53.00 & 0.00 / 0.00 / 0.00 \\
            VCTREE \cite{tang2019learning} & 48.38 / 55.69 / 58.05 & 2.76 / 3.47 / 3.95 & 42.96 / 52.28 / 55.75 & 0.00 / 0.00 / 0.00 \\
            TDE \cite{tang2020unbiased} & 52.49 / 60.18 / 62.47 & 2.68 / 3.49 / 4.01 &  48.90 / 58.69 / 61.50 & 0.00 / 0.00 / 0.00 \\
            $\text{VS}^3$ \cite{zhang2023learning} (\footnotesize Swin-T)& 39.83 / 46.73 / 49.11 & 9.85 / 14.49 / 17.87 &
            36.16 / 44.51 / 47.63  & 6.02 / 10.24 / 13.44 \\    
            \hline 
            
            {OvSGTR} (\footnotesize Swin-T) &  {54.05 / 60.58 / 62.10} & {12.34 / 18.14 / 23.20} & {51.47 / 59.01 / 60.65}& {6.90 / 12.06 / 16.49} \\
            
           {OvSGTR} (\footnotesize Swin-B) &  53.81 / 59.83 / 61.34 &   \textbf{15.43} / \textbf{21.35} / \textbf{26.22}   &  51.16 / 58.30 / 60.00&    10.21 / 15.58 / 19.96 \\
             
          $\text{OvSGTR}^{\star}$(\footnotesize Swin-T) &
          52.24 / 58.60 / 60.35 &   14.33 / 20.91 / 25.98 
           & 50.67 / 58.18 / 60.15 & \textbf{10.52} / \textbf{17.30} / \textbf{22.90} 
          \\ 
           
          $\text{OvSGTR}^{\star}$(\footnotesize Swin-B) &  \textbf{54.74} / \textbf{60.93} / \textbf{62.49} &    15.21 / 21.21 / 26.12  &  \textbf{53.21} / \textbf{60.78} / \textbf{62.60} &   10.31 / 15.78 / 20.47  \\            
          \bottomrule
    \end{tabular}}
    \label{tab:ovd}
\end{table*}
\subsubsection{Closed-set SGG Benchmark} 
The \textit{closed-set SGG} setting follows previous works~\cite{zellers2018neural,tang2020unbiased,li2022sgtr,xu2017scene,zhang2023learning} and employs the VG150 dataset~\cite{xu2017scene} with full manual annotations for training and evaluation. 
Experimental results on the VG150 test set (Table~\ref{tab:closed}) show that our proposed model outperforms all competitors. 
Notably, compared to the recent $\text{VS}^3$~\cite{zhang2023learning}, our \textit{OvSGTR} (with Swin-T) achieves performance gains of up to $4.9\%$ for R@50 and $6.9\%$ for R@100. 
Improvements in mR@K further indicate that our model better addresses long-tail bias. 
Moreover, the performance improvement (with vs. without relation-aware pretraining, \emph{e.g.}, 27.8/36.4/42.4 vs. 28.6/37.6/43.4) demonstrates that large-scale relation-aware pre-training is beneficial for recognizing both visual relationships and objects.

While many prior approaches rely on complex message-passing mechanisms to extract relation features, our model attains strong performance with a simpler relation head consisting of only two MLP layers. 
For instance, \textit{OvSGTR} (with Swin-T) achieves results comparable to, or even superior to, those of $\text{VS}^3$ (with Swin-L), 
while requiring fewer trainable parameters (41M vs. 124M) and exhibiting lower inference latency (0.13\,s vs. 0.24\,s).
\begin{table*}[t]
    \centering
    \caption{Experimental results of \textit{OvR-SGG} setting on VG150 test set.
    $\ddagger$ denotes \textit{w.o.} relation-aware pre-training.  
    $\dag$ denotes \textit{w.o.} distillation but \textit{w.} relation-aware pre-training on COCO Caption (\ie, using the Scene Parser-based Pipeline).
    }    
    \resizebox{0.985\textwidth}{!}{
    \begin{tabular}{l|>{\centering\arraybackslash}p{24mm}>{\centering\arraybackslash}p{24mm}|>{\centering\arraybackslash}p{24mm}>{\centering\arraybackslash}p{24mm}
    } 
            \toprule
            \multirow{2}{*}{Method} & \multicolumn{2}{c|}{Base+Novel (Relation)} & 
           \multicolumn{2}{c}{Novel (Relation)}  \\  
              & PredCls &  SGDet &   PredCls & SGDet 
              \\ 
          \midrule 
          IMP \cite{xu2017scene} &   23.41 / 25.74 / 26.59  & 9.16 / 12.42 / 14.48 & 0.00 / 0.00 / 0.00 &  0.00 / 0.00 / 0.00    \\
          MOTIFS \cite{zellers2018neural} &24.82 / 27.13 / 27.99  & 12.41 / 15.29 / 16.81&  0.00 / 0.00 / 0.00&   0.00 / 0.00 / 0.00\\
          VCTREE \cite{tang2019learning} & 26.91 / 29.34 / 30.16 & 12.38 / 15.11 / 16.81 & 0.00 / 0.00 / 0.00  & 0.00 / 0.00 / 0.00  \\
          TDE \cite{tang2020unbiased} &26.69 / 28.98 / 29.76 &  12.43 / 15.37 / 17.21 &  0.00 / 0.00 / 0.00  &  0.00 / 0.00 / 0.00 \\ 
        $\text{VS}^3$ \cite{zhang2023learning} (\footnotesize Swin-T) &
          22.27 / 24.32 / 24.99 & 12.45 / 15.63 / 17.29 & 0.00 / 0.00 / 0.00  & 0.00 / 0.00 / 0.00  \\ 
        \hline 
       $\text{OvSGTR (\footnotesize Swin-T)}^\ddagger$  &  26.57 / 28.46 / 29.09   & 14.60 / 18.09 / 20.41  &  0.00 / 0.00 / 0.00 & 0.00 / 0.00 / 0.00  \\ 
       $\text{OvSGTR (\footnotesize Swin-T)}^\dag$  &  25.57 / 27.63 / 28.28   & 14.32 / 17.69 / 19.96  &   0.17 / 0.25 / 0.27 & 0.05 / 0.10 / 0.15 \\ 
        OvSGTR (\footnotesize Swin-T)  &  31.83 / 37.58 / 40.68  &  15.85 / 20.46 / 23.86 &  21.01 / 27.42 / 31.30 &  10.17 / 13.45 / 16.19 \\ 
           $\text{OvSGTR}^{\star}$ (\footnotesize Swin-T) &
        40.60 / 46.82 / 49.03
        &19.38 / 25.40 / 29.71 & 
        28.91 / 36.78 / 39.96 & 
        12.23 / 17.02 / 21.15 
        \\  \hdashline      
        $\text{OvSGTR (\footnotesize Swin-B)}^\ddagger$  & 26.83 / 28.83 / 29.42   & 15.39 / 19.07 / 21.37   & 0.00 / 0.00 / 0.00 & 0.00 / 0.00 / 0.00  \\ 
        $\text{OvSGTR (\footnotesize Swin-B)}^\dag$  &  26.21 / 28.30 / 28.96  &  15.00 / 18.58 / 20.84   & 0.10 / 0.13 / 0.14  & 0.06 / 0.08 / 0.10  \\         
        OvSGTR (\footnotesize Swin-B) &  34.37 / 41.04 / 44.67   &  17.63 / 22.89 / 26.65  &  24.95 / 32.85 / 37.87  &   12.09 / 16.39 / 19.72  \\         
       $\text{OvSGTR}^{\star}$ (\footnotesize Swin-B) & \textbf{44.53} / \textbf{51.31} / \textbf{53.71}  & \textbf{21.09} / \textbf{27.92} / \textbf{32.74}  &  
       \textbf{38.27} / \textbf{47.42} / \textbf{50.90} & 
       \textbf{16.59} / \textbf{22.86} / \textbf{27.73}  \\                
          \bottomrule
    \end{tabular}}
    \label{tab:ovr}
\end{table*}
\begin{table*}[t]
    \centering
    \caption{Experimental results of \textit{OvD+R-SGG} setting on VG150 test set. 
    $\dag$ denotes \textit{w.o.} distillation but \textit{w.} relation-aware pre-training on COCO Caption (\ie, using the Scene Parser-based Pipeline).
    }    
    \resizebox{0.985\textwidth}{!}
    {
    \begin{tabular}{l|ccc|ccc|ccc} 
            \toprule
            \multirow{2}{*}{Method}  &\multicolumn{3}{c|}{Joint Base+Novel} & 
           \multicolumn{3}{c}{Novel (Object)}  & \multicolumn{3}{c}{Novel (Relation)}  \\
              & R@20 & R@50  & R@100  &   R@20 &  R@50  & R@100  &    R@20 & R@50 & R@100 \\ 
          \midrule 
          IMP \cite{xu2017scene} &  0.57 & 0.81  &  0.99 & 0.00 & 0.00 & 0.00 & 0.00 & 0.00 & 0.00 \\
          MOTIFS \cite{zellers2018neural} &0.77 & 0.95 & 1.10 & 0.00 & 0.00 & 0.00 & 0.00 & 0.00 & 0.00 \\
          VCTREE \cite{tang2019learning} & 0.84 &1.06 & 1.17 & 0.00 & 0.00 &0.00 & 0.00 & 0.00 & 0.00\\
          TDE \cite{tang2020unbiased} &0.81 &0.99 & 1.13 &0.00 & 0.00 & 0.00 & 0.00 & 0.00 &0.00\\         
        $\text{VS}^3$ \cite{zhang2023learning} (\footnotesize Swin-T) &   4.03 &5.87 & 7.19 &  3.73 & 5.98 & 7.47 & 0.00 & 0.00 & 0.00 \\ 
        \hline 
        
         $\text{OvSGTR (\footnotesize Swin-T)}^\dag$ & 5.20 & 7.88   &  10.06  & 4.18 & 6.82  &9.23  &0.00 &0.00 &0.00 \\ 
        
        OvSGTR (\footnotesize Swin-T) & 10.02  & 13.50 & 16.37 &  10.56 & 14.32   & 17.48  & 7.09 & 9.19 & 11.18\\ 
           $\text{OvSGTR}^{\star}$ (\footnotesize Swin-T) & 10.67
& 15.15  &18.82  & 8.22 & 12.49    & 16.29 & 9.62 &13.68  & 17.19  \\    \hdashline       
       $\text{OvSGTR (\footnotesize Swin-B)}^\dag$ &  7.85 & 11.23 & 14.21     &9.26  & 13.27  &  16.83  &1.20 &  1.78 & 2.57 \\   
        OvSGTR (\footnotesize Swin-B) & 12.37  &17.14  & 21.03   & \textbf{12.63} & \textbf{17.58}    &  \textbf{21.70}   & 10.56 & 14.62   & 18.22  \\  
        $\text{OvSGTR}^{\star}$ (\footnotesize Swin-B) &  \textbf{12.54}
&\textbf{17.84}  &\textbf{21.95}  & {10.29} &{15.66}    & {19.84}   & \textbf{12.21} &\textbf{17.15}  & \textbf{21.05}  \\ 
          \bottomrule
    \end{tabular}}
    \label{tab:ovdr}
\end{table*}
\subsubsection{OvD-SGG Benchmark} 
Following previous works~\cite{he2022towards, zhang2023learning}, the \textit{OvD-SGG} setting ensures that models do not see novel object categories during training. 
Specifically, $70\%$ of the VG150 object categories are designated as base categories, and the remaining $30\%$ serve as novel categories. 
The experiments mirror those in the \textit{Closed-set SGG} scenario, 
except that annotations for novel object categories are omitted during training. 
After excluding unseen object nodes, the VG150 training set contains $50{,}107$ images.

Table~\ref{tab:ovd} reports the performance under the \textit{OvD-SGG} setting in terms of ``Base+Novel (Object)'' and ``Novel (Object).'' 
The results show that our proposed model substantially outperforms previous methods, indicating a strong open-vocabulary recognition and generalization capability. 
Compared to $\text{VS}^3$~\cite{zhang2023learning}, our approach yields improvements of up to $68.9\%$ and $70.4\%$ in R@50 and R@100, respectively, on novel categories. 
Since \textit{OvD-SGG} only excludes nodes of novel object categories, the learning process for relations remains unaffected. 
This highlights that the open-vocabulary ability of the object detector plays a more decisive role in \textit{OvD-SGG}.

\subsubsection{OvR-SGG Benchmark.} 
Different from \textit{OvD-SGG} which removes all unseen nodes, 
\textit{OvR-SGG} only removes unseen edges but retains the original nodes. 
Considering that VG150 has $50$ relation categories, we randomly select $15$ of them as unseen (novel) relation categories. 
During training, only base relation annotations are available. 
After removing these unseen edges, $44{,}333$ images from VG150 remain for training.

Similar to \textit{OvD-SGG}, Table~\ref{tab:ovr} reports the performance of \textit{OvR-SGG} in terms of both ``Base+Novel (Relation)'' and ``Novel (Relation).'' 
From Table~\ref{tab:ovr}, the proposed \textit{OvSGTR} notably outperforms other competitors, even without distillation. 
However, a marked decline in performance is observed across all methods (including \textit{OvSGTR} without distillation) in the ``Novel (Relation)'' categories, 
highlighting the inherent difficulty of recognizing novel relations in the \textit{OvR-SGG} paradigm. 
Nevertheless, with visual-concept retention, 
the performance of \textit{OvSGTR} (\textit{w/} Swin-T) on novel relations is significantly improved from $0.10$ (R@50, SGDet) to $13.45$ (R@50, SGDet).

Further, with large-scale relation-aware pre-training on MegaSG, 
\emph{OvSGTR} (\textit{w/} Swin-B) achieves a notable performance gain 
(\eg, $16.59 / 22.86 / 27.37$ vs.\ $12.09 / 16.39 / 19.72$ in SGDet for novel relations) 
compared to pre-training on COCO Caption. 
This result showcases that scaling relation-aware pre-training is both valuable and effective.
\subsubsection{OvD+R-SGG Benchmark.}
To further extend the SGG task to a fully open-vocabulary domain, 
we propose the \textit{OvD+R-SGG} benchmark, 
where both novel object and novel relation categories are excluded during training. 
Specifically, we merge the splits from \textit{OvD-SGG} and \textit{OvR-SGG} using their respective base object and base relation categories, 
resulting in $36{,}425$ training images from VG150.

Table~\ref{tab:ovdr} presents the performance of \textit{OvD+R-SGG} under three settings: \textit{Joint Base+Novel} (\ie, all object and relation categories), \textit{Novel (Object)} (\ie, only novel object categories), and \textit{Novel (Relation)} (\ie, only novel relation categories). 
As with \textit{OvR-SGG}, we observe catastrophic forgetting in \textit{OvD+R-SGG}, 
but our proposed visual-concept retention significantly mitigates this issue. 
Compared to existing methods, our model achieves notable gains across all metrics. 
For instance, under the \textit{Joint Base+Novel} setting, 
the proposed \emph{OvSGTR} (w/ Swin-B) achieves $17.84\%$ (R@50, SGDet), 
surpassing $\text{VS}^3$ (5.87\%) by $11.97$ points. 
Moreover, for \textit{Novel (Object)} categories, 
our approach attains $15.66\%$ (R@50, SGDet, compared to 5.98\% from $\text{VS}^3$), and for \textit{Novel (Relation)} categories, 
\emph{OvSGTR} (w/ Swin-B) reaches $17.15\%$ (R@50, SGDet), outperforming $\text{VS}^3$, which failed to recognize novel relationships.
\subsubsection{Overall Analysis}  
Experimental results highlight distinct challenges across the four settings. 
Based on these findings:  
\textbf{1)} Many previous methods rely on a two-stage object detector, Faster R-CNN \cite{ren2015faster}, and complex message-passing mechanisms. 
However, our model demonstrates that a one-stage DETR-based framework can significantly outperform R-CNN-like architectures, even when using only a single MLP for relation feature representation.  
\textbf{2)} Previous methods employing a closed-set object detector struggle to recognize objects without textual cues in object-involved open-vocabulary SGG (\emph{i.e.}, \textit{OvD-SGG} and \textit{OvD+R-SGG}).  
\textbf{3)} The performance drop in \textit{OvD+R-SGG} compared to other settings suggests it is significantly more challenging, underscoring the need for further exploration toward fully open-vocabulary SGG.  
\textbf{4)} Large-scale relation-aware pre-training substantially enhances the generalization ability of SGG models across both closed-set and open-vocabulary scenarios. 
This indicates that the synthesized SGG dataset plays a crucial role in bridging the gap between these settings.  
\subsection{Ablation Study}
\begin{figure}[t]
\centering
\definecolor{cbBlue}{RGB}{0,107,164} 
\definecolor{cbRed}{RGB}{200,82,0} 
\definecolor{cbGreen}{RGB}{16,115,0} 
\begin{tikzpicture}
\begin{axis}[
    ybar,
    x=14mm, 
    width=0.5\textwidth,
    height=40mm, 
    bar width=10pt,
    symbolic x coords={0,1,3,5},
    xtick=data,
    ylabel style={font=\footnotesize, scale=0.8},
    xlabel style={font=\footnotesize, scale=0.8}, 
    x tick label style={align=center, font=\footnotesize}, 
    y tick label style={font=\footnotesize}, 
    nodes near coords,
    nodes near coords align={vertical},
    every node near coord/.append style={font=\footnotesize, color=black, scale=0.65}, 
    ymin=10, ymax=45,
    legend style={
        at={(0.5, 1.32)},
        anchor=north,
        legend columns=-1,
        font=\footnotesize,
        scale=0.8
    },
    enlarge x limits={0.2}, 
]

\addplot[fill=cbBlue] coordinates {(0, 25.44) (1, 25.73) (3, 24.84) (5, 25.13)};
\addplot[fill=cbRed] coordinates {(0, 33.45) (1, 33.92) (3, 33.24) (5, 33.49)};
\addplot[fill=cbGreen!70] coordinates {(0, 39.12) (1, 39.48) (3, 38.94) (5, 39.23)};

\legend{R@20,R@50,R@100}
\end{axis}
\end{tikzpicture}
\caption{Ablation study of  relation queries on VG150 validation set under the setting of \textit{Closed-set SGG}.}
\label{fig:ablation1}
\end{figure}
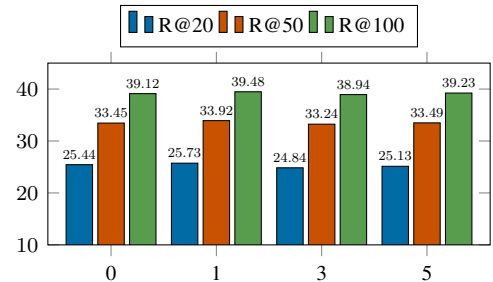 

\textit{Effect of Relation Queries.}
We first consider removing the relation query embedding. In this case, the relation feature is computed by 
\begin{equation}
    \bm{e}_{s_i \rightarrow o_j} = f_{\theta}\bigl([\bm{v}_{s_i}, \bm{v}_{o_j}]\bigr),
\end{equation}
which only encodes features from the subject and object nodes. We further extend Eq.~\eqref{eq:rln} to a more general form:
\begin{equation}
    \bm{e}_{s_i \rightarrow o_j} = \frac{1}{M}\sum_{n=1}^M f_{\theta}\bigl([\bm{v}_{s_i}, \bm{v}_{o_j}, \bm{r}_n]\bigr),
    \label{eq:avg@rln}
\end{equation}
where multiple relation queries are averaged. As shown in Fig.~\ref{fig:ablation1}, the model achieves the highest performance when the number of relation queries is set to 1. This result can be interpreted from two perspectives: (\romannumeral1) relation queries interact with all edges during training, thereby capturing global information from the entire dataset; (\romannumeral2) increasing the number of relation-aware queries does not introduce additional supervision but increases the optimization burden.

\textit{Hyper-parameter $\lambda$ for Distillation.}
Table~\ref{tab:lambda} shows the impact of varying the hyper-parameter $\lambda$. When $\lambda=0.1$, the model achieves the best performance. By contrast, removing distillation leads to a substantial drop in performance for novel categories, indicating that the model struggles to retain the knowledge inherited from pre-trained models for these categories.

\begin{table}[t]
    \centering
    \caption{Impact of hyper-parameter $\lambda$ for distillation loss on VG150 validation set under the setting of \textit{OvR-SGG}.
    $a \rightarrow b$ refers to the performance shift from $a$ (initial checkpoint's performance) to $b$ during training. }    
    \resizebox{\columnwidth}{!}
    {
    \begin{tabular}{c|cc|cc}
        \toprule
         $\lambda$  & \multicolumn{2}{c|}{Base+Novel} &
          \multicolumn{2}{c}{Novel} \\
          &  R@50 &  R@100 & R@50 &  R@100 \\
        \midrule
         0 & 7.25 $\rightarrow$ 13.74  & 8.98  $\rightarrow$ 16.11
         & 10.78 $\rightarrow$ 0.32 & 13.24 $\rightarrow$ 0.38 \\ 
         0.1 &   7.25 $\rightarrow$ \textbf{16.00}  & 8.98  $\rightarrow$ \textbf{19.20}
         & 10.78 $\rightarrow$ \textbf{11.54} & 13.24 $\rightarrow$ \textbf{13.94} \\ 
         0.3 &     7.25 $\rightarrow$ 14.35  & 8.98 $\rightarrow$ 17.04  &
         10.78 $\rightarrow$  10.71 & 13.24 $\rightarrow$  12.71
         \\
        0.5 &     7.25 $\rightarrow$ 13.34 & 8.98 $\rightarrow$ 16.08 &
         10.78 $\rightarrow$ 10.90 & 13.24 $\rightarrow$ 13.22 
         \\
       \bottomrule
    \end{tabular}}
    \label{tab:lambda}
\end{table}

\input{pics/fig_vis}
\subsection{Extension: Comparison on GQA Benchmark}
\begin{table}[t]
    \centering
\caption{Experimental results on the GQA200 test set. 
OvSGTR is pre-trained on MegaSG. 
``OvSGTR-T'' refers to our model with the Swin-T backbone, 
while ``OvSGTR-B'' denotes the variant with the Swin-B backbone.}   
    \resizebox{\columnwidth}{!}{%
    \begin{tabular}{l|cc|cc} 
        \toprule
        \multirow{2}{*}{Method} & \multicolumn{2}{c|}{PredCls} & \multicolumn{2}{c}{SGDet} \\ 
        & R@50 & R@100 & R@50 & R@100 \\ 
        \midrule 
            IMP \cite{xu2017scene} & 57.5 & 59.6 &  22.3&  26.7\\ 
            MOTIFS \cite{zellers2018neural}& 60.2 & 62.1 & 24.8 & 28.8\\
            VCTREE \cite{tang2019learning}&62.2
            & 63.9 & 27.1 & 30.8\\
            TDE \cite{tang2020unbiased} & 64.2 & 66.0 &  27.2 & 31.5 \\ 
        \midrule
         OvSGTR-T (\emph{w/o} fine-tuning) &   38.0 &  40.5  & 11.6  &   13.9\\ 
        OvSGTR-T (\emph{w/} fine-tuning)  &  60.7 & 62.7   & 32.8 &  37.9 \\ 
          OvSGTR-B(\emph{w/o} fine-tuning) &  38.2   & 40.9   & 11.4  &   13.9\\ 
        OvSGTR-B (\emph{w/} fine-tuning)  &  61.5  &  63.4   &  \textbf{33.9}  &  \textbf{39.2}  \\        
        \bottomrule
    \end{tabular}%
    }
    \label{tab:gqa} 
\end{table}
Beyond the widely used VG150 benchmark, we also validate our framework on the GQA dataset.
Following prior works~\cite{dong2022stacked,sudhakaran2023vision,li2024leveraging}, 
we use the GQA200 split that contains $200$ objects and $100$ predicates. 
As shown in Table~\ref{tab:gqa}, our proposed \emph{OvSGTR} demonstrates a strong generalization capability on GQA200.
Without fine-tuning, both Swin-T and Swin-B variants of \emph{OvSGTR} achieve remarkable performance in the PredCls setting. However, their SGDet scores remain low due to domain shift and lack of localization supervision.

After fine-tuning the GQA200 training set, both variants show significant improvements, particularly under the SGDet protocol.
Notably, OvSGTR-B achieves {33.9\%} (R@50) and {39.2\%} (R@100) in SGDet, surpassing all previous competitors.
These results highlight the effectiveness of our relation-aware pre-training and visual-concept alignment mechanisms,
which enable the model to adapt to novel domains with minimal supervision.

Overall, the GQA results further validate the scalability and effectiveness of \emph{OvSGTR} under different dataset distributions,
reinforcing its applicability in real-world, diverse visual environments.
\subsection{Visualization and Discussion}
\subsubsection{Relation-aware pre-training}

 \definecolor{colorArrow2}{RGB}{173, 216, 230}

 \tikzset{
  myarrow2/.style={
line width=3pt, -{Triangle[length=2mm, width=2mm]}, color=colorArrow2
  },
}
\begin{figure*}[t]
    \centering
    \resizebox{\textwidth}{!}{
    \begin{tikzpicture}
        \node (a) at (0, 0) {\includegraphics[width=0.2\textwidth]{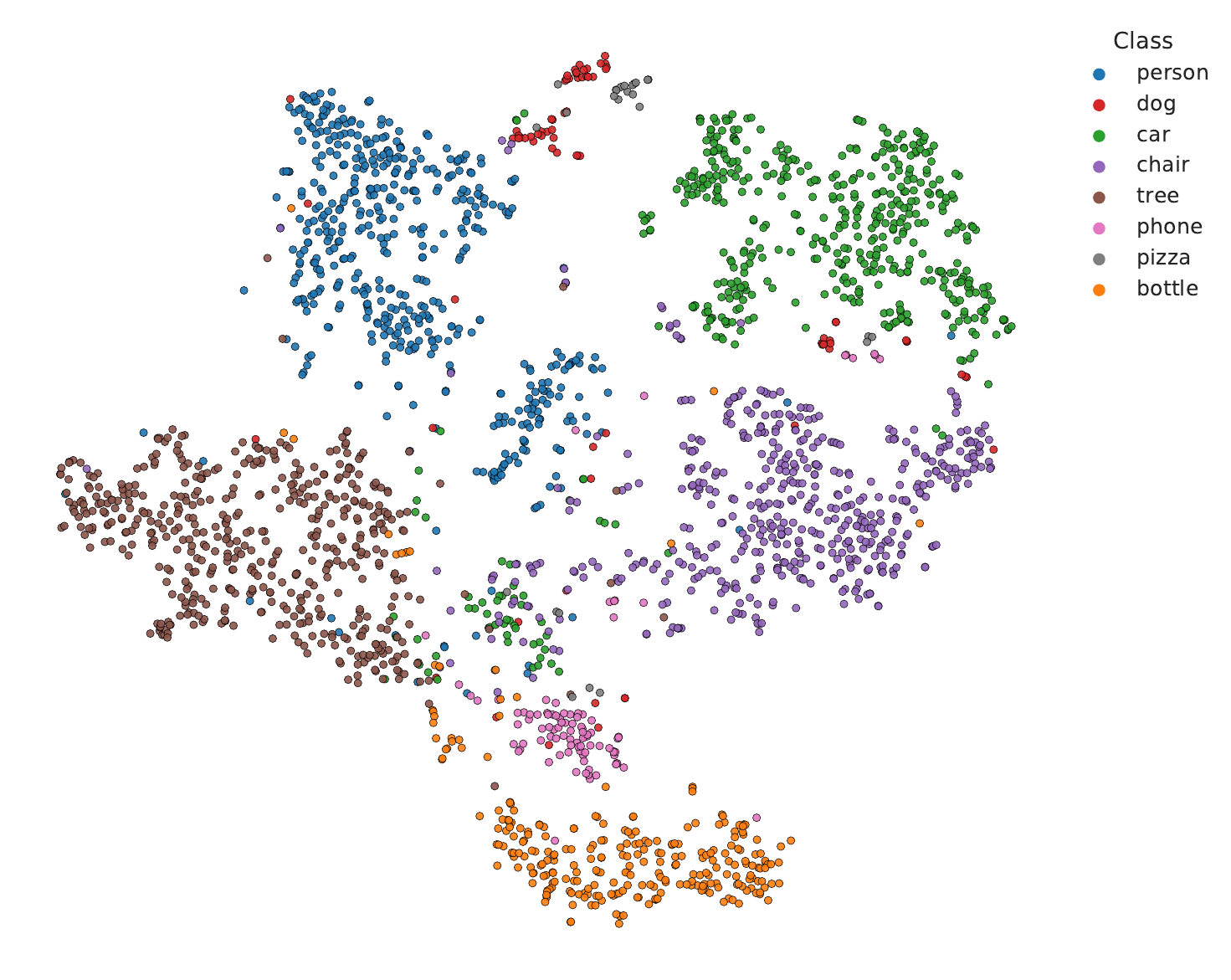}};
        \node (b) at (4.2, 0) {\includegraphics[width=0.2\textwidth]{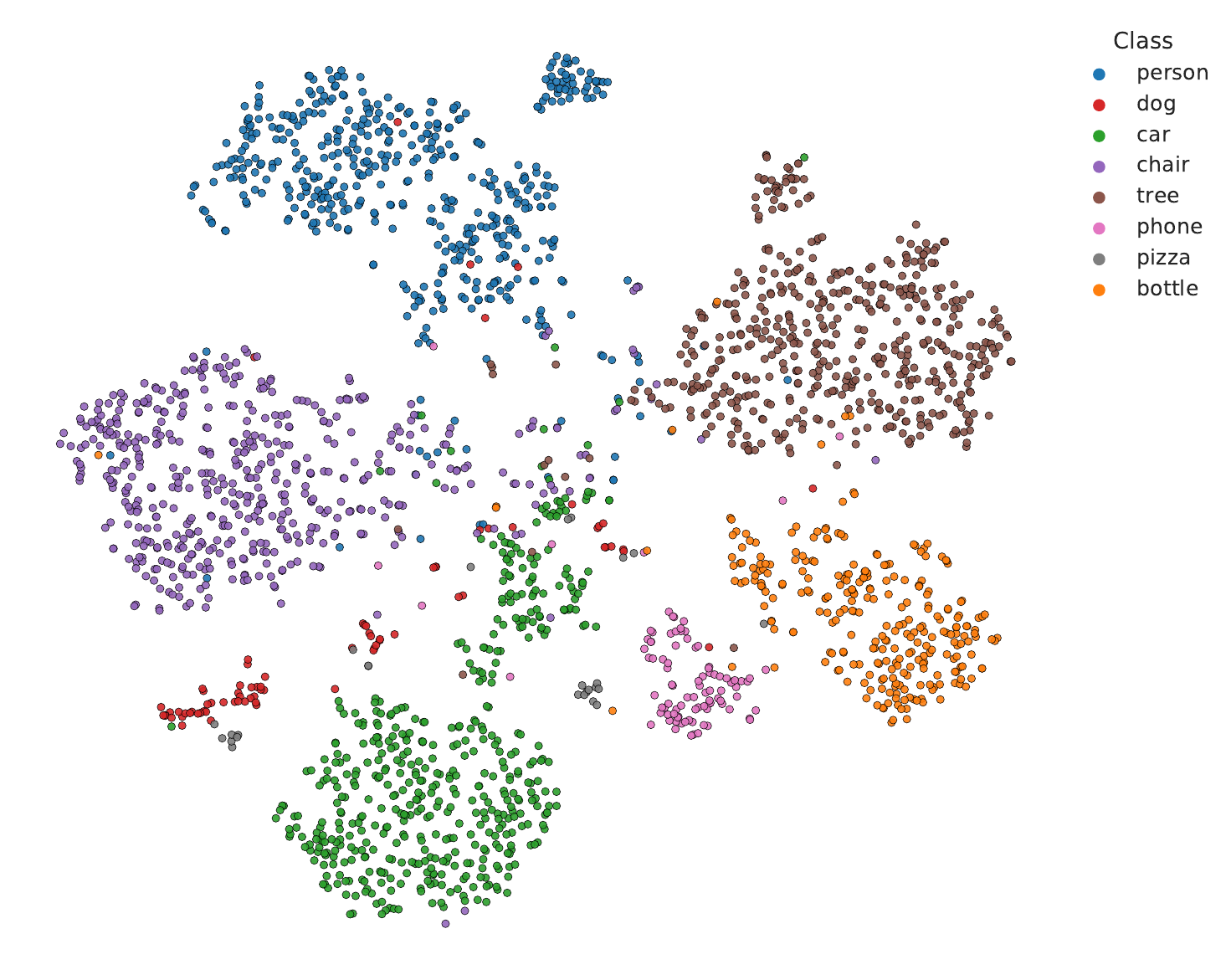}};
\path (a) -- (b) coordinate[midway] (abMid);
\node[below=15mm of abMid] {\footnotesize (a) Object features (\emph{w/o} relation-aware pre-training).};        
        \node (c) at (8.4, 0) {\includegraphics[width=0.2\textwidth]{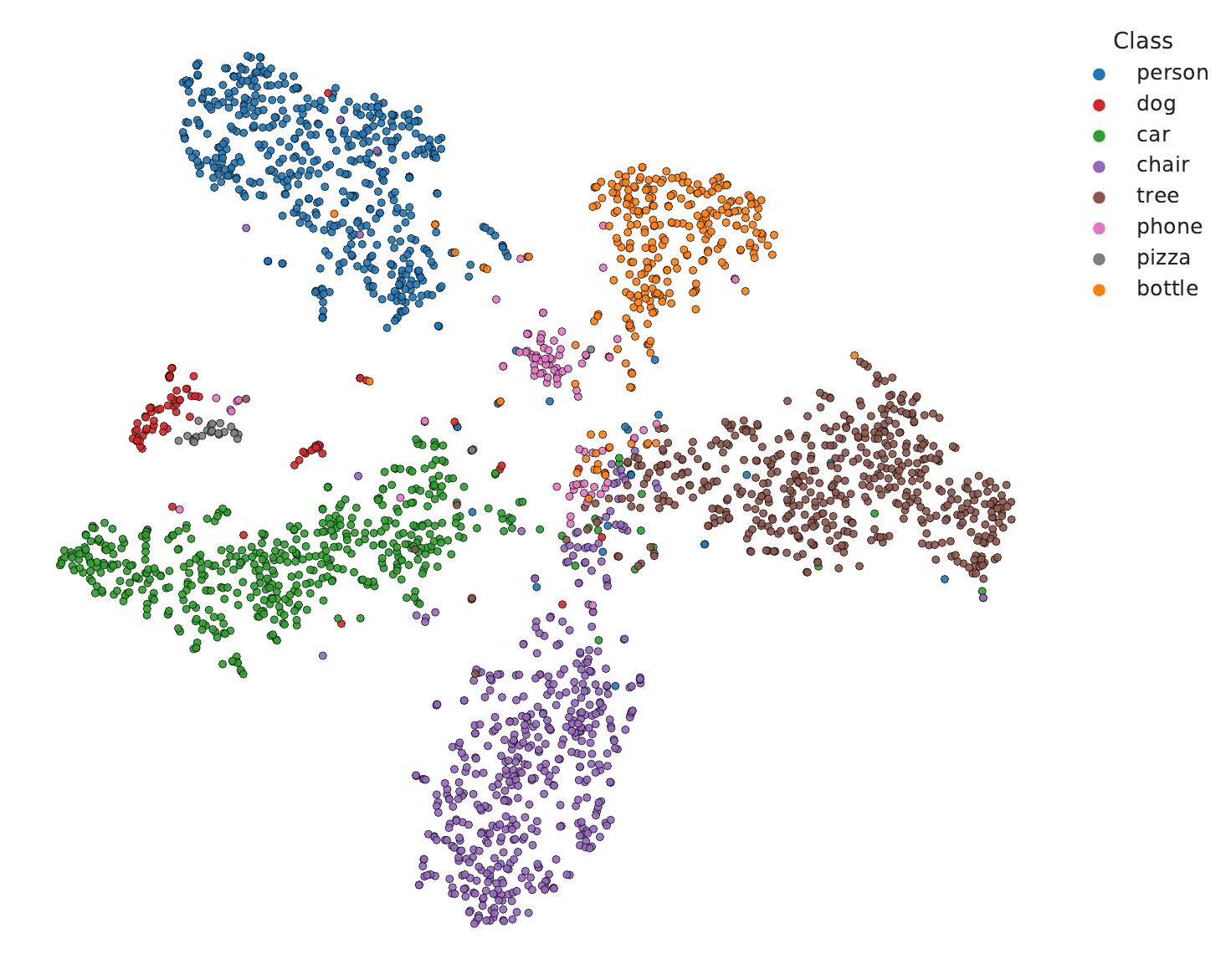}};
        \node (d) at (12.6, 0) {\includegraphics[width=0.2\textwidth]{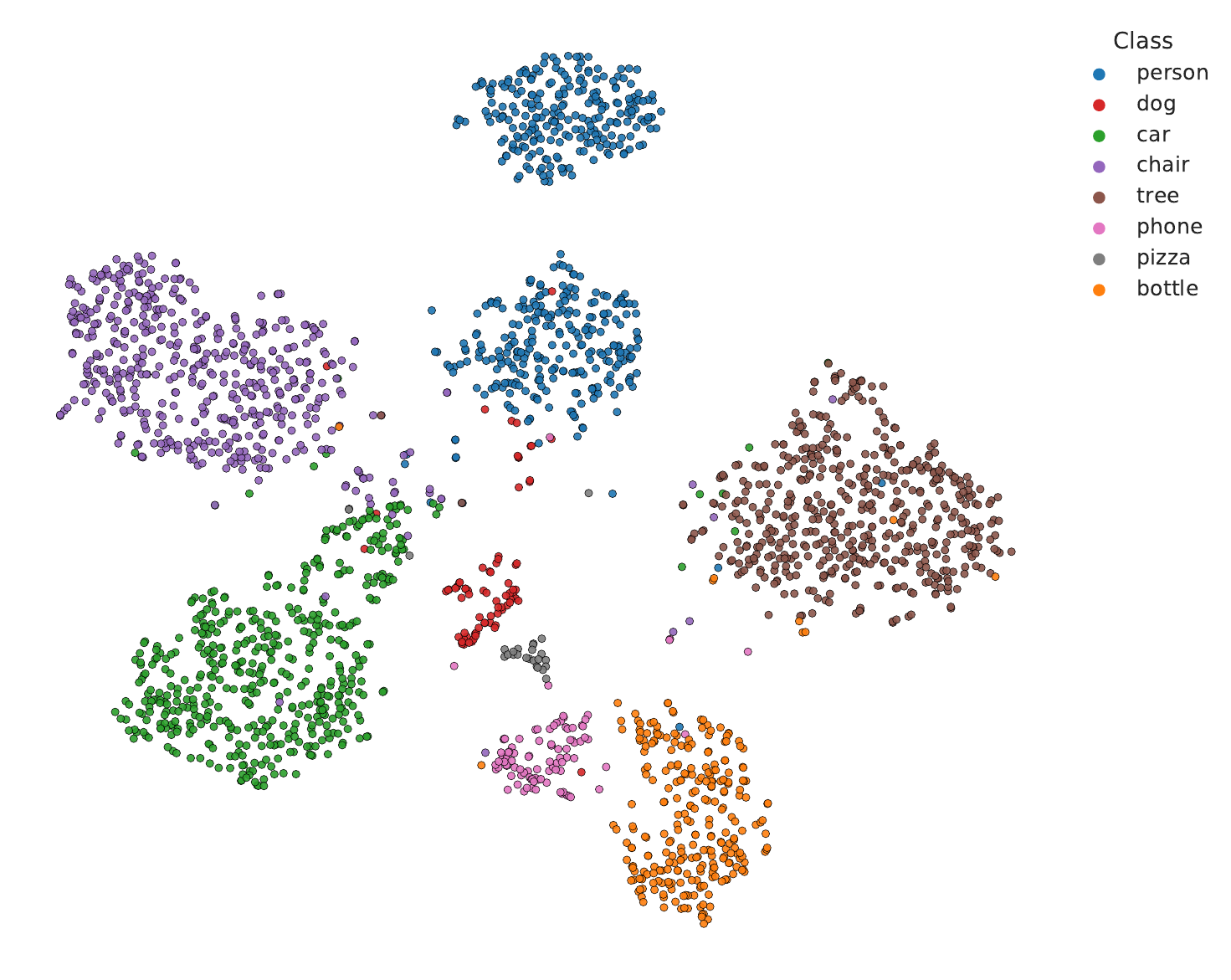}};
\path (c) -- (d) coordinate[midway] (cdMid);
\node[below=15mm of cdMid] {\footnotesize (b) Object features (\emph{w/} relation-aware pre-training).}; 
        \node (e) at (0, -4.2) {\includegraphics[width=0.2\textwidth]{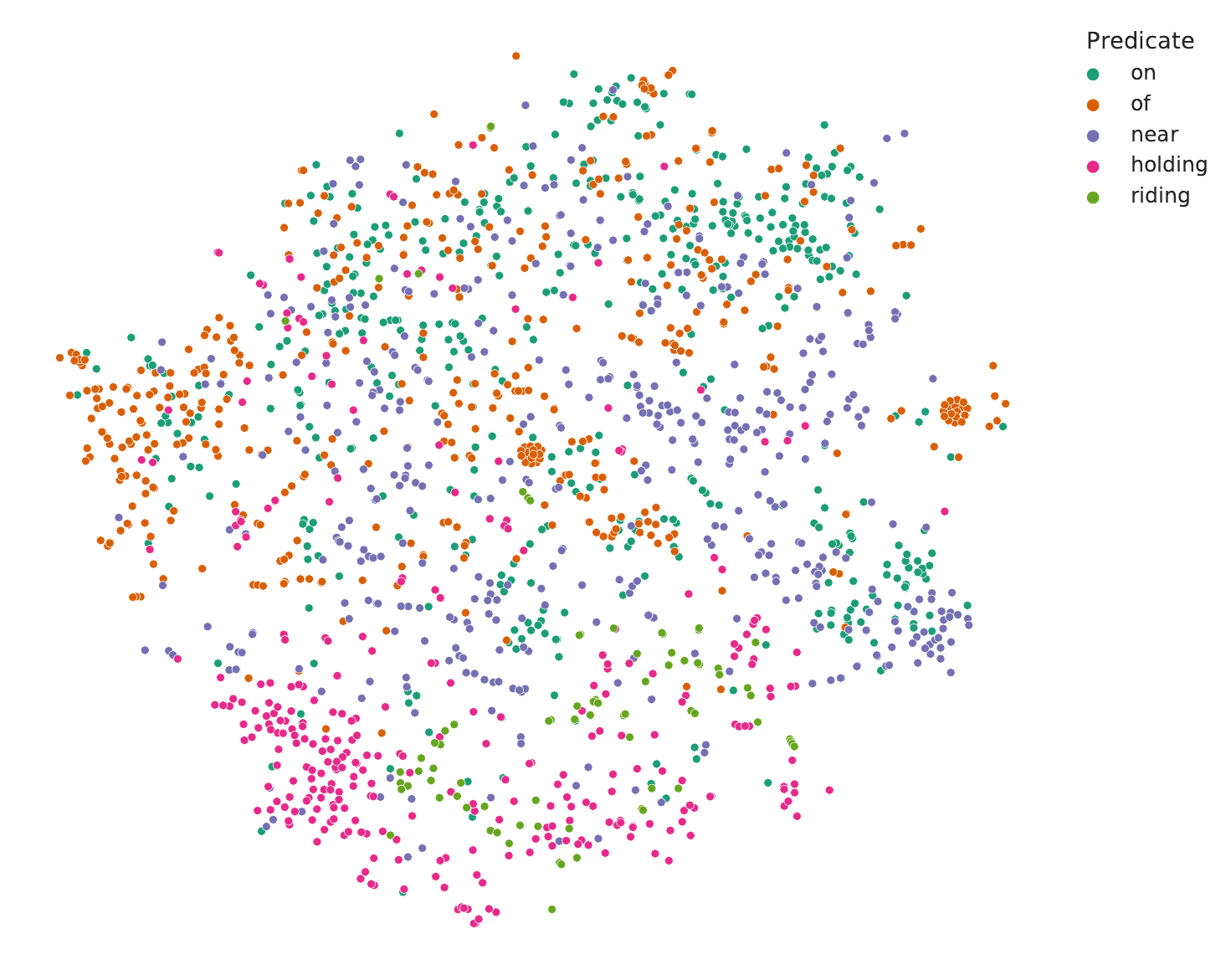}};
        \node (f) at (4.2, -4.2) {\includegraphics[width=0.2\textwidth]{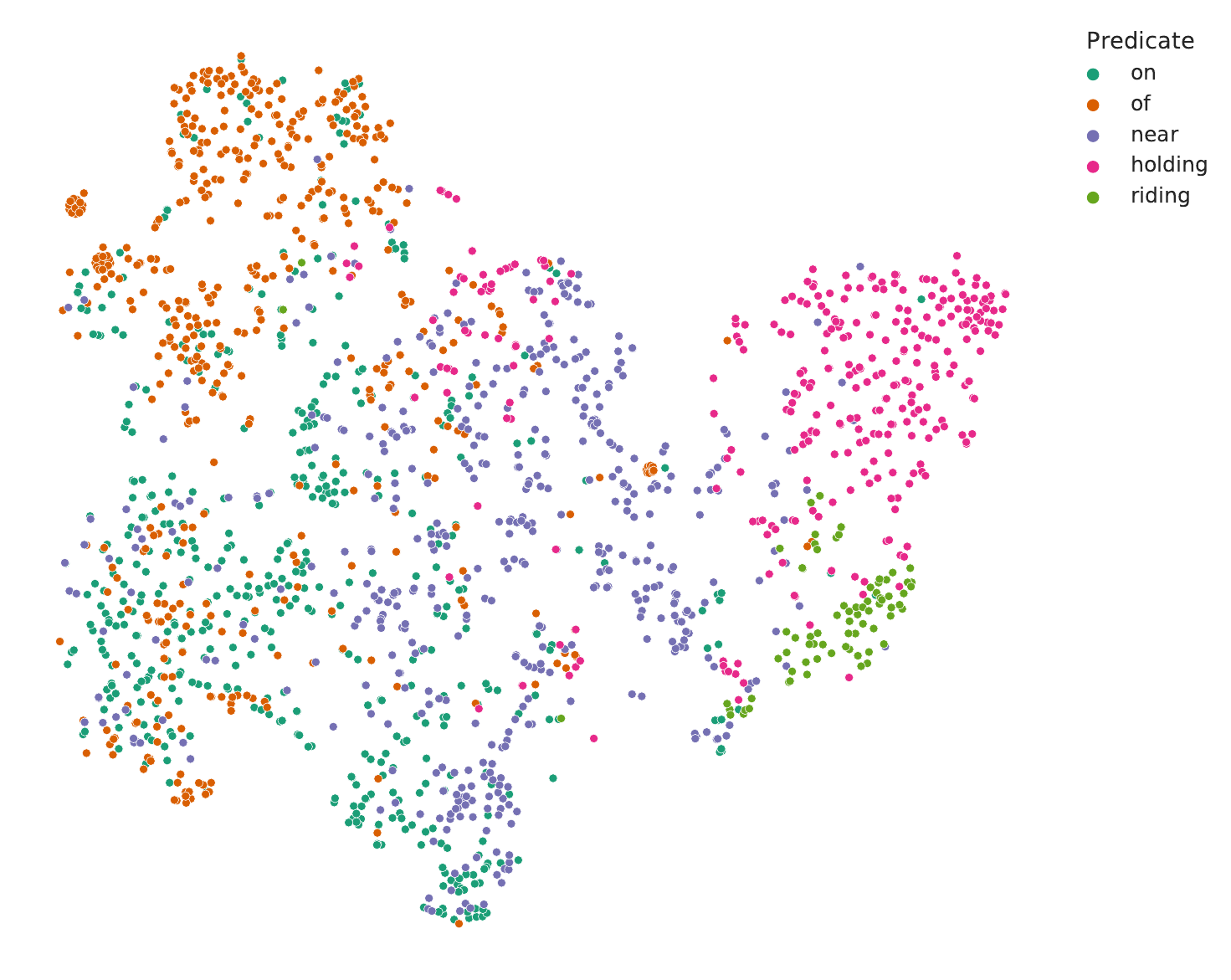}};
\path (e) -- (f) coordinate[midway] (efMid);
\node[below=15mm of efMid] {\footnotesize (c) Relation features (\emph{w/o} relation-aware pre-training).};         
        \node (g) at (8.4, -4.2) {\includegraphics[width=0.2\textwidth]{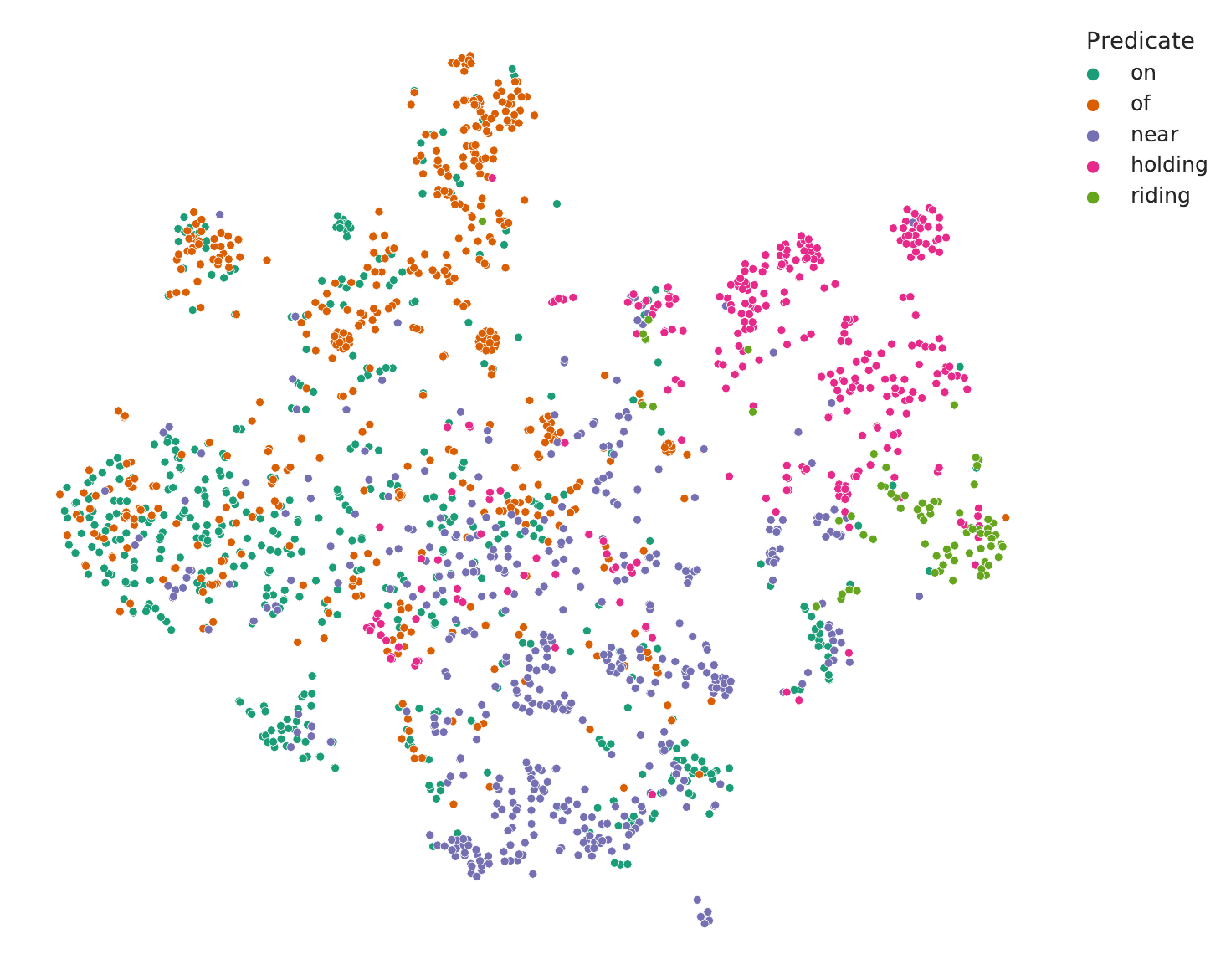}};
        \node (h) at (12.6, -4.2) {\includegraphics[width=0.2\textwidth]{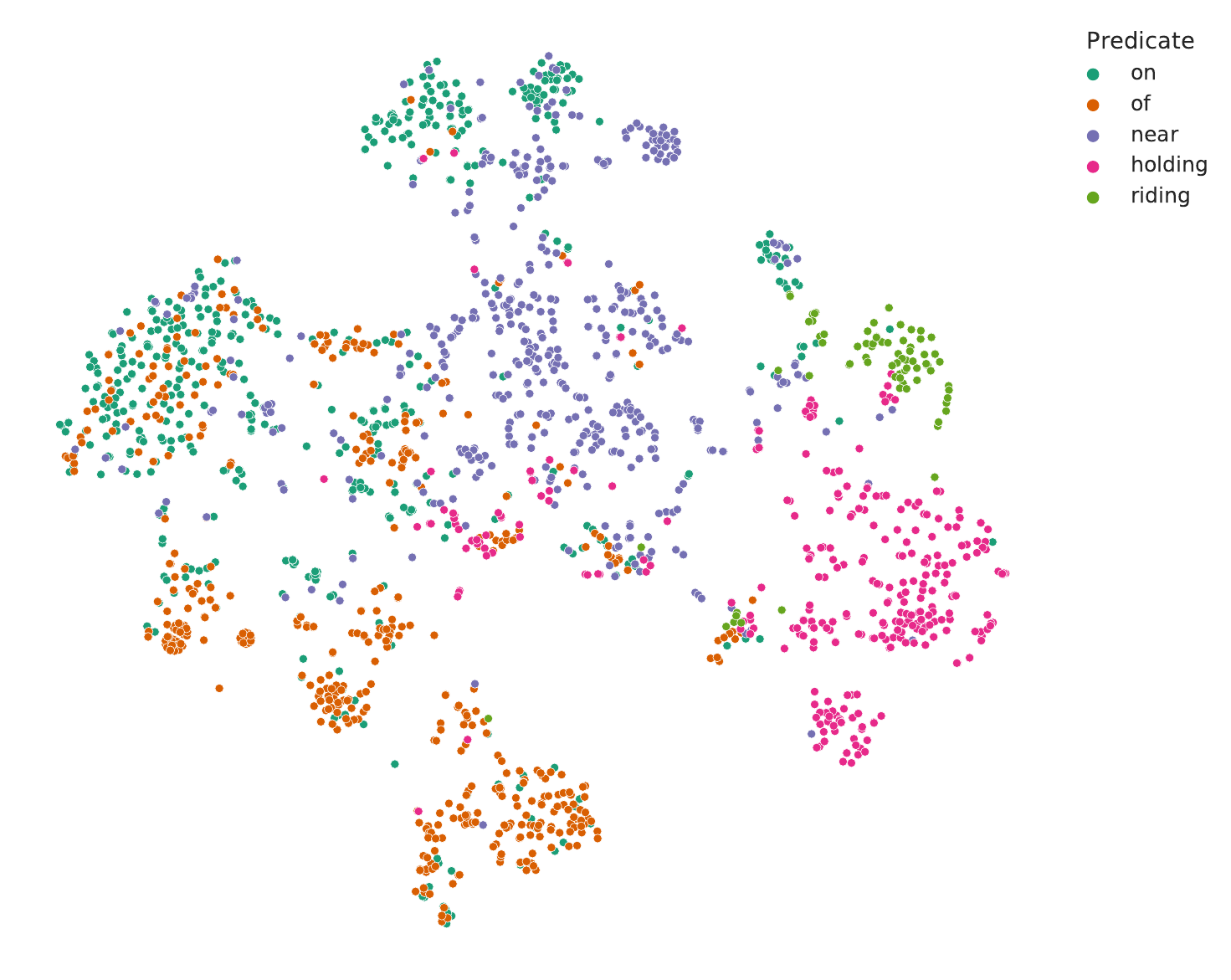}};
\path (g) -- (h) coordinate[midway] (ghMid);
\node[below=15mm of ghMid] {\footnotesize (d) Relation features (\emph{w/} relation-aware pre-training).};   
        
\draw[myarrow2] (a.east) -- (b.west);  
\draw[myarrow2] (c.east) -- (d.west);  
\draw[myarrow2] (e.east) -- (f.west);  
\draw[myarrow2] (g.east) -- (h.west);  

    \end{tikzpicture}
}
\caption{t-SNE~\cite{van2008visualizing} visualizations of object and relation features. 
(a)–(b) show object features extracted by OvSGTR-B without and with relation-aware pre-training, respectively. 
(c)–(d) show relation features under the same settings. Arrows indicate the evolution of features after fine-tuning on VG150.
}
    \label{fig:tsne-feature}
\end{figure*}
To enrich the understanding of relation-aware pre-training, 
we visualize the object and relation features using t-SNE~\cite{van2008visualizing} in Fig.~\ref{fig:tsne-feature}. 
Object and relation features become more discriminative and semantically clustered after pre-training on \emph{MegaSG}.
For instance, object features like ``dog'', ``pizza'', and ``phone'' form tighter, more coherent clusters. 
Similarly, relation features such as ``on,'' ``riding'', and ``holding'' become more separable, 
resulting in better classification of relationships.

This relation-aware pre-training significantly boosts generalization, especially in open-vocabulary scenarios, where models must reason over novel object and relation categories. 
By learning transferable representations from weak supervision, our framework achieves superior zero-shot performance and mitigates reliance on fully annotated datasets.
\subsubsection{Qualitative Results}
We present qualitative results of our model trained under four settings, as shown in Fig.~\ref{fig:qual}.
From the figure, the model trained on the \textit{Closed-set SGG} setting tends to generate denser scene graphs, as all object and relationship categories are available during training.
In contrast, the model trained under the \textit{OvD-SGG} setting demonstrates the ability to detect novel objects (\eg, ``bat'', ``bus''), although the relationships are limited to the seen predicate set.
Under the \textit{OvR-SGG} setting, all object nodes are retained, but the model must infer novel relationships. 
This results in sparser graphs; however, the model still predicts some unseen relations (\eg, ``on'', ``of'') with reasonable accuracy.
Notably, the model trained on \textit{OvD+R-SGG}, despite lacking full supervision of both novel objects and relationships, is still able to recognize unseen object categories such as ``bus'' and ``bat'' (which does not exist in the VG150 dataset), 
along with novel relationships like ``on''. 
This highlights the strong generalization ability of our approach under fully open-vocabulary settings.

\section{Conclusion} 
This work advances scene graph generation (SGG) from a closed-set paradigm to a fully open-vocabulary setting by considering both object nodes and relationship edges. 
We categorize SGG scenarios into four distinct settings—\emph{Closed-SGG}, \emph{OvD-SGG}, \emph{OvR-SGG}, and \emph{OvD+R-SGG}—and introduce \emph{OvSGTR}, a unified transformer-based framework that addresses these challenges. 
Our framework aligns visual features with semantic information from both base and novel categories, 
enabling it to capture complex inter-object relationships and generalize to unseen objects and relations.
To obtain a transferable representation for relations, 
we studied three relation-aware pre-training pipelines and adopted a knowledge distillation strategy to prevent catastrophic forgetting during fine-tuning. 
Extensive experiments on the VG150 benchmark demonstrate that our method achieves new state-of-the-art performance across all settings. 
These results highlight the benefits of integrating large-scale relation-aware pre-training with effective relation-aware strategies, contributing to improved scene graph generation.  
\par
\ifCLASSOPTIONcaptionsoff
  \newpage
\fi
{
\bibliographystyle{IEEEtran}
\bibliography{ref}

\begin{thebibliography}{10}
\providecommand{\url}[1]{#1}
\csname url@samestyle\endcsname
\providecommand{\newblock}{\relax}
\providecommand{\bibinfo}[2]{#2}
\providecommand{\BIBentrySTDinterwordspacing}{\spaceskip=0pt\relax}
\providecommand{\BIBentryALTinterwordstretchfactor}{4}
\providecommand{\BIBentryALTinterwordspacing}{\spaceskip=\fontdimen2\font plus
\BIBentryALTinterwordstretchfactor\fontdimen3\font minus \fontdimen4\font\relax}
\providecommand{\BIBforeignlanguage}[2]{{%
\expandafter\ifx\csname l@#1\endcsname\relax
\typeout{** WARNING: IEEEtran.bst: No hyphenation pattern has been}%
\typeout{** loaded for the language `#1'. Using the pattern for}%
\typeout{** the default language instead.}%
\else
\language=\csname l@#1\endcsname
\fi
#2}}
\providecommand{\BIBdecl}{\relax}
\BIBdecl

\bibitem{xu2017scene}
D.~Xu, Y.~Zhu, C.~B. Choy, and L.~Fei{-}Fei, ``Scene graph generation by iterative message passing,'' in \emph{IEEE Conf. Comput. Vis. Pattern Recog.}, 2017, pp. 3097--3106.

\bibitem{zellers2018neural}
R.~Zellers, M.~Yatskar, S.~Thomson, and Y.~Choi, ``Neural motifs: Scene graph parsing with global context,'' in \emph{IEEE Conf. Comput. Vis. Pattern Recog.}, 2018, pp. 5831--5840.

\bibitem{tang2019learning}
K.~Tang, H.~Zhang, B.~Wu, W.~Luo, and W.~Liu, ``Learning to compose dynamic tree structures for visual contexts,'' in \emph{IEEE Conf. Comput. Vis. Pattern Recog.}, 2019, pp. 6619--6628.

\bibitem{tang2020unbiased}
K.~Tang, Y.~Niu, J.~Huang, J.~Shi, and H.~Zhang, ``Unbiased scene graph generation from biased training,'' in \emph{IEEE Conf. Comput. Vis. Pattern Recog.}, 2020, pp. 3713--3722.

\bibitem{chiou2021recovering}
M.~Chiou, H.~Ding, H.~Yan, C.~Wang, R.~Zimmermann, and J.~Feng, ``Recovering the unbiased scene graphs from the biased ones,'' in \emph{ACM Int. Conf. Multimedia}, 2021, pp. 1581--1590.

\bibitem{li2021bipartite}
R.~Li, S.~Zhang, B.~Wan, and X.~He, ``Bipartite graph network with adaptive message passing for unbiased scene graph generation,'' in \emph{IEEE Conf. Comput. Vis. Pattern Recog.}, 2021, pp. 11\,109--11\,119.

\bibitem{chen2019knowledge}
T.~Chen, W.~Yu, R.~Chen, and L.~Lin, ``Knowledge-embedded routing network for scene graph generation,'' in \emph{IEEE Conf. Comput. Vis. Pattern Recog.}, 2019, pp. 6163--6171.

\bibitem{zhang2019graphical}
J.~Zhang, K.~J. Shih, A.~Elgammal, A.~Tao, and B.~Catanzaro, ``Graphical contrastive losses for scene graph parsing,'' in \emph{IEEE Conf. Comput. Vis. Pattern Recog.}, 2019, pp. 11\,535--11\,543.

\bibitem{he2022towards}
T.~He, L.~Gao, J.~Song, and Y.~Li, ``Towards open-vocabulary scene graph generation with prompt-based finetuning,'' in \emph{Eur. Conf. Comput. Vis.}, 2022, pp. 56--73.

\bibitem{zhang2023learning}
Y.~Zhang, Y.~Pan, T.~Yao, R.~Huang, T.~Mei, and C.~W. Chen, ``Learning to generate language-supervised and open-vocabulary scene graph using pre-trained visual-semantic space,'' in \emph{IEEE Conf. Comput. Vis. Pattern Recog.}, 2023, pp. 2915--2924.

\bibitem{yang2019auto}
X.~Yang, K.~Tang, H.~Zhang, and J.~Cai, ``Auto-encoding scene graphs for image captioning,'' in \emph{IEEE Conf. Comput. Vis. Pattern Recog.}, 2019, pp. 10\,685--10\,694.

\bibitem{chen2020say}
S.~Chen, Q.~Jin, P.~Wang, and Q.~Wu, ``Say as you wish: Fine-grained control of image caption generation with abstract scene graphs,'' in \emph{IEEE Conf. Comput. Vis. Pattern Recog.}, 2020, pp. 9959--9968.

\bibitem{gu2019unpaired}
J.~Gu, S.~R. Joty, J.~Cai, H.~Zhao, X.~Yang, and G.~Wang, ``Unpaired image captioning via scene graph alignments,'' in \emph{Int. Conf. Comput. Vis.}, 2019, pp. 10\,322--10\,331.

\bibitem{wang2019role}
D.~Wang, D.~Beck, and T.~Cohn, ``On the role of scene graphs in image captioning,'' in \emph{LANTERN@EMNLP-IJCNLP}, 2019, pp. 29--34.

\bibitem{nguyen2021defense}
K.~Nguyen, S.~Tripathi, B.~Du, T.~Guha, and T.~Q. Nguyen, ``In defense of scene graphs for image captioning,'' in \emph{Int. Conf. Comput. Vis.}, 2021, pp. 1387--1396.

\bibitem{teney2017graph}
D.~Teney, L.~Liu, and A.~van~den Hengel, ``Graph-structured representations for visual question answering,'' in \emph{IEEE Conf. Comput. Vis. Pattern Recog.}, 2017, pp. 3233--3241.

\bibitem{nuthalapati2021lightweight}
S.~V. Nuthalapati, R.~Chandradevan, E.~Giunchiglia, B.~Li, M.~Kayser, T.~Lukasiewicz, and C.~Yang, ``Lightweight visual question answering using scene graphs,'' in \emph{CIKM}, 2021, pp. 3353--3357.

\bibitem{kenfack2020robotvqa}
F.~K. Kenfack, F.~A. Siddiky, F.~Balint{-}Benczedi, and M.~Beetz, ``Robotvqa - {A} scene-graph- and deep-learning-based visual question answering system for robot manipulation,'' in \emph{IROS}, 2020, pp. 9667--9674.

\bibitem{lee2019visual}
S.~Lee, J.~Kim, Y.~Oh, and J.~H. Jeon, ``Visual question answering over scene graph,'' in \emph{Proc. Int. Conf. Graph Comput.}, 2019, pp. 45--50.

\bibitem{johnson2018image}
J.~Johnson, A.~Gupta, and L.~Fei{-}Fei, ``Image generation from scene graphs,'' in \emph{IEEE Conf. Comput. Vis. Pattern Recog.}, 2018, pp. 1219--1228.

\bibitem{yang2022diffusion}
L.~Yang, Z.~Huang, Y.~Song, S.~Hong, G.~Li, W.~Zhang, B.~Cui, B.~Ghanem, and M.~Yang, ``Diffusion-based scene graph to image generation with masked contrastive pre-training,'' \emph{CoRR}, vol. abs/2211.11138, 2022.

\bibitem{werby2024hierarchical}
A.~Werby, C.~Huang, M.~B{\"{u}}chner, A.~Valada, and W.~Burgard, ``Hierarchical open-vocabulary 3d scene graphs for language-grounded robot navigation,'' \emph{CoRR}, vol. abs/2403.17846, 2024.

\bibitem{yin2024sg}
H.~Yin, X.~Xu, Z.~Wu, J.~Zhou, and J.~Lu, ``Sg-nav: Online 3d scene graph prompting for llm-based zero-shot object navigation,'' \emph{CoRR}, vol. abs/2410.08189, 2024.

\bibitem{miao2024sgloc}
Y.~Miao, F.~Engelmann, O.~Vysotska, F.~Tombari, M.~Pollefeys, and D.~B. Bar{\'{a}}th, ``Scenegraphloc: Cross-modal coarse visual localization on 3d scene graphs,'' in \emph{Eur. Conf. Comput. Vis.}, A.~Leonardis, E.~Ricci, S.~Roth, O.~Russakovsky, T.~Sattler, and G.~Varol, Eds., vol. 15066, 2024, pp. 127--150.

\bibitem{zhong2021learning}
Y.~Zhong, J.~Shi, J.~Yang, C.~Xu, and Y.~Li, ``Learning to generate scene graph from natural language supervision,'' in \emph{Int. Conf. Comput. Vis.}, 2021, pp. 1823--1834.

\bibitem{li2022integrating}
X.~Li, L.~Chen, W.~Ma, Y.~Yang, and J.~Xiao, ``Integrating object-aware and interaction-aware knowledge for weakly supervised scene graph generation,'' in \emph{ACM Int. Conf. Multimedia}, 2022, pp. 4204--4213.

\bibitem{chen2023gpt4sgg}
Z.~Chen, J.~Wu, Z.~Lei, Z.~Zhang, and C.~Chen, ``{GPT4SGG}: Synthesizing scene graphs from holistic and region-specific narratives,'' \emph{arXiv preprint arXiv:2312.04314}, 2023.

\bibitem{zareian2021open}
A.~Zareian, K.~D. Rosa, D.~H. Hu, and S.~Chang, ``Open-vocabulary object detection using captions,'' in \emph{IEEE Conf. Comput. Vis. Pattern Recog.}, 2021, pp. 14\,393--14\,402.

\bibitem{wu2023aligning}
S.~Wu, W.~Zhang, S.~Jin, W.~Liu, and C.~C. Loy, ``Aligning bag of regions for open-vocabulary object detection,'' in \emph{IEEE Conf. Comput. Vis. Pattern Recog.}, 2023, pp. 15\,254--15\,264.

\bibitem{li2022grounded}
L.~H. Li, P.~Zhang, H.~Zhang, J.~Yang, C.~Li, Y.~Zhong, L.~Wang, L.~Yuan, L.~Zhang, J.~Hwang, K.~Chang, and J.~Gao, ``Grounded language-image pre-training,'' in \emph{IEEE Conf. Comput. Vis. Pattern Recog.}, 2022, pp. 10\,955--10\,965.

\bibitem{zhong2022regionclip}
Y.~Zhong, J.~Yang, P.~Zhang, C.~Li, N.~Codella, L.~H. Li, L.~Zhou, X.~Dai, L.~Yuan, Y.~Li, and J.~Gao, ``Regionclip: Region-based language-image pretraining,'' in \emph{IEEE Conf. Comput. Vis. Pattern Recog.}, 2022, pp. 16\,772--16\,782.

\bibitem{du2022learning}
Y.~Du, F.~Wei, Z.~Zhang, M.~Shi, Y.~Gao, and G.~Li, ``Learning to prompt for open-vocabulary object detection with vision-language model,'' in \emph{IEEE Conf. Comput. Vis. Pattern Recog.}, 2022, pp. 14\,064--14\,073.

\bibitem{mao2018parser}
J.~Mao, ``Scene graph parser,'' \url{https : / / github . com / vacancy / SceneGraphParser}, 2022.

\bibitem{DBLP:conf/iccv/LiOZWW17}
Y.~Li, W.~Ouyang, B.~Zhou, K.~Wang, and X.~Wang, ``Scene graph generation from objects, phrases and region captions,'' in \emph{Int. Conf. Comput. Vis.}, 2017, pp. 1270--1279.

\bibitem{DBLP:conf/cvpr/ChenYCL19}
T.~Chen, W.~Yu, R.~Chen, and L.~Lin, ``Knowledge-embedded routing network for scene graph generation,'' in \emph{IEEE Conf. Comput. Vis. Pattern Recog.}, 2019, pp. 6163--6171.

\bibitem{li2022sgtr}
R.~Li, S.~Zhang, and X.~He, ``Sgtr: End-to-end scene graph generation with transformer,'' in \emph{IEEE Conf. Comput. Vis. Pattern Recog.}, 2022, pp. 19\,464--19\,474.

\bibitem{DBLP:conf/nips/KhandelwalS22}
S.~Khandelwal and L.~Sigal, ``Iterative scene graph generation,'' in \emph{Adv. Neural Inform. Process. Syst.}, S.~Koyejo, S.~Mohamed, A.~Agarwal, D.~Belgrave, K.~Cho, and A.~Oh, Eds., 2022.

\bibitem{DBLP:journals/pami/CongYR23}
Y.~Cong, M.~Y. Yang, and B.~Rosenhahn, ``Reltr: Relation transformer for scene graph generation,'' \emph{{IEEE} Trans. Pattern Anal. Mach. Intell.}, vol.~45, no.~9, pp. 11\,169--11\,183, 2023.

\bibitem{DBLP:journals/pami/SunZLHL23}
S.~Sun, S.~Zhi, Q.~Liao, J.~Heikkil{\"{a}}, and L.~Liu, ``Unbiased scene graph generation via two-stage causal modeling,'' \emph{{IEEE} Trans. Pattern Anal. Mach. Intell.}, vol.~45, no.~10, pp. 12\,562--12\,580, 2023.

\bibitem{DBLP:conf/cvpr/JinGMZXWMS23}
T.~Jin, F.~Guo, Q.~Meng, S.~Zhu, X.~Xi, W.~Wang, Z.~Mu, and W.~Song, ``Fast contextual scene graph generation with unbiased context augmentation,'' in \emph{IEEE Conf. Comput. Vis. Pattern Recog.}, 2023, pp. 6302--6311.

\bibitem{DBLP:conf/eccv/JeonKYP24}
J.~Jeon, K.~Kim, K.~Yoon, and C.~Park, ``Semantic diversity-aware prototype-based learning for unbiased scene graph generation,'' in \emph{Eur. Conf. Comput. Vis.}, A.~Leonardis, E.~Ricci, S.~Roth, O.~Russakovsky, T.~Sattler, and G.~Varol, Eds., vol. 15126, 2024, pp. 379--395.

\bibitem{ren2015faster}
S.~Ren, K.~He, R.~Girshick, and J.~Sun, ``Faster r-cnn: Towards real-time object detection with region proposal networks,'' \emph{Adv. Neural Inform. Process. Syst.}, vol.~28, 2015.

\bibitem{chen2024makes}
Z.~Chen, J.~Wu, Z.~Lei, and C.~W. Chen, ``What makes a scene? scene graph-based evaluation and feedback for controllable generation,'' \emph{arXiv preprint arXiv:2411.15435}, 2024.

\bibitem{radford2021learning}
A.~Radford, J.~W. Kim, C.~Hallacy, A.~Ramesh, G.~Goh, S.~Agarwal, G.~Sastry, A.~Askell, P.~Mishkin, J.~Clark, G.~Krueger, and I.~Sutskever, ``Learning transferable visual models from natural language supervision,'' in \emph{ICML}, 2021, pp. 8748--8763.

\bibitem{liu2023grounding}
S.~Liu, Z.~Zeng, T.~Ren, F.~Li, H.~Zhang, J.~Yang, C.~Li, J.~Yang, H.~Su, J.~Zhu, and L.~Zhang, ``Grounding {DINO:} marrying {DINO} with grounded pre-training for open-set object detection,'' \emph{CoRR}, vol. abs/2303.05499, 2023.

\bibitem{chen2015microsoft}
X.~Chen, H.~Fang, T.~Lin, R.~Vedantam, S.~Gupta, P.~Doll{\'{a}}r, and C.~L. Zitnick, ``Microsoft {COCO} captions: Data collection and evaluation server,'' \emph{CoRR}, vol. abs/1504.00325, 2015.

\bibitem{DBLP:conf/iclr/GuLKC22}
X.~Gu, T.~Lin, W.~Kuo, and Y.~Cui, ``Open-vocabulary object detection via vision and language knowledge distillation,'' in \emph{Int. Conf. Learn. Represent.}, 2022.

\bibitem{DBLP:conf/eccv/GhiasiGCL22}
G.~Ghiasi, X.~Gu, Y.~Cui, and T.~Lin, ``Scaling open-vocabulary image segmentation with image-level labels,'' in \emph{Eur. Conf. Comput. Vis.}, 2022, pp. 540--557.

\bibitem{DBLP:journals/corr/abs-2109-08472}
M.~Wang, J.~Xing, and Y.~Liu, ``Actionclip: {A} new paradigm for video action recognition,'' \emph{CoRR}, vol. abs/2109.08472, 2021.

\bibitem{li2024pixels}
R.~Li, S.~Zhang, D.~Lin, K.~Chen, and X.~He, ``From pixels to graphs: Open-vocabulary scene graph generation with vision-language models,'' in \emph{IEEE Conf. Comput. Vis. Pattern Recog.}, 2024, pp. 28\,076--28\,086.

\bibitem{wu2024towards}
J.~Wu, X.~Li, S.~Xu, H.~Yuan, H.~Ding, Y.~Yang, X.~Li, J.~Zhang, Y.~Tong, X.~Jiang \emph{et~al.}, ``Towards open vocabulary learning: A survey,'' \emph{IEEE Trans. Pattern Anal. Mach. Intell.}, 2024.

\bibitem{DBLP:journals/corr/abs-2307-09220}
C.~Zhu and L.~Chen, ``A survey on open-vocabulary detection and segmentation: Past, present, and future,'' \emph{CoRR}, vol. abs/2307.09220, 2023.

\bibitem{kim2024llm4sgg}
K.~Kim, K.~Yoon, J.~Jeon, Y.~In, J.~Moon, D.~Kim, and C.~Park, ``{LLM4SGG}: Large language models for weakly supervised scene graph generation,'' in \emph{IEEE Conf. Comput. Vis. Pattern Recog.}, 2024, pp. 28\,306--28\,316.

\bibitem{OpenAI2023GPT4V}
OpenAI, ``{GPT-4v(ision) System Card},'' \url{https://openai.com/research/gpt-4v-system-card}, 2023.

\bibitem{chen2024expanding}
Z.~Chen, J.~Wu, Z.~Lei, Z.~Zhang, and C.~W. Chen, ``Expanding scene graph boundaries: fully open-vocabulary scene graph generation via visual-concept alignment and retention,'' in \emph{Eur. Conf. Comput. Vis.}, 2024, pp. 108--124.

\bibitem{liu2021swin}
Z.~Liu, Y.~Lin, Y.~Cao, H.~Hu, Y.~Wei, Z.~Zhang, S.~Lin, and B.~Guo, ``Swin transformer: Hierarchical vision transformer using shifted windows,'' in \emph{Int. Conf. Comput. Vis.}, 2021, pp. 9992--10\,002.

\bibitem{kenton2019bert}
J.~Devlin, M.~Chang, K.~Lee, and K.~Toutanova, ``{BERT:} pre-training of deep bidirectional transformers for language understanding,'' in \emph{NAACL-HLT}, 2019, pp. 4171--4186.

\bibitem{zhu2021deformable}
X.~Zhu, W.~Su, L.~Lu, B.~Li, X.~Wang, and J.~Dai, ``Deformable {DETR:} deformable transformers for end-to-end object detection,'' in \emph{Int. Conf. Learn. Represent.}, 2021.

\bibitem{rezatofighi2019generalized}
H.~Rezatofighi, N.~Tsoi, J.~Gwak, A.~Sadeghian, I.~D. Reid, and S.~Savarese, ``Generalized intersection over union: {A} metric and a loss for bounding box regression,'' in \emph{IEEE Conf. Comput. Vis. Pattern Recog.}, 2019, pp. 658--666.

\bibitem{lin2017focal}
T.~Lin, P.~Goyal, R.~B. Girshick, K.~He, and P.~Doll{\'{a}}r, ``Focal loss for dense object detection,'' in \emph{Int. Conf. Comput. Vis.}, 2017, pp. 2999--3007.

\bibitem{achiam2023gpt}
J.~Achiam, S.~Adler, S.~Agarwal, L.~Ahmad, I.~Akkaya, F.~L. Aleman, D.~Almeida, J.~Altenschmidt, S.~Altman, S.~Anadkat \emph{et~al.}, ``Gpt-4 technical report,'' \emph{arXiv preprint arXiv:2303.08774}, 2023.

\bibitem{reid2024gemini}
M.~Reid, N.~Savinov, D.~Teplyashin, D.~Lepikhin, T.~Lillicrap, J.-b. Alayrac, R.~Soricut, A.~Lazaridou, O.~Firat, J.~Schrittwieser \emph{et~al.}, ``Gemini 1.5: Unlocking multimodal understanding across millions of tokens of context,'' \emph{arXiv preprint arXiv:2403.05530}, 2024.

\bibitem{hudson2019gqa}
D.~A. Hudson and C.~D. Manning, ``Gqa: A new dataset for real-world visual reasoning and compositional question answering,'' in \emph{IEEE Conf. Comput. Vis. Pattern Recog.}, 2019, pp. 6700--6709.

\bibitem{krishna2017visual}
R.~Krishna, Y.~Zhu, O.~Groth, J.~Johnson, K.~Hata, J.~Kravitz, S.~Chen, Y.~Kalantidis, L.-J. Li, D.~A. Shamma \emph{et~al.}, ``Visual genome: Connecting language and vision using crowdsourced dense image annotations,'' \emph{IJCV}, vol. 123, pp. 32--73, 2017.

\bibitem{dong2022stacked}
X.~Dong, T.~Gan, X.~Song, J.~Wu, Y.~Cheng, and L.~Nie, ``Stacked hybrid-attention and group collaborative learning for unbiased scene graph generation,'' in \emph{IEEE Conf. Comput. Vis. Pattern Recog.}, 2022, pp. 19\,427--19\,436.

\bibitem{sudhakaran2023vision}
G.~Sudhakaran, D.~S. Dhami, K.~Kersting, and S.~Roth, ``Vision relation transformer for unbiased scene graph generation,'' in \emph{Int. Conf. Comput. Vis.}, 2023, pp. 21\,882--21\,893.

\bibitem{adamw}
I.~Loshchilov and F.~Hutter, ``Decoupled weight decay regularization,'' in \emph{Int. Conf. Learn. Represent.}, 2019.

\bibitem{yelinguistic}
K.~Ye and A.~Kovashka, ``Linguistic structures as weak supervision for visual scene graph generation,'' in \emph{IEEE Conf. Comput. Vis. Pattern Recog.}, 2021, pp. 8289--8299.

\bibitem{chen2020uniter}
Y.~Chen, L.~Li, L.~Yu, A.~E. Kholy, F.~Ahmed, Z.~Gan, Y.~Cheng, and J.~Liu, ``{UNITER:} universal image-text representation learning,'' in \emph{Eur. Conf. Comput. Vis.}, 2020, pp. 104--120.

\bibitem{lin2022hl}
X.~Lin, C.~Ding, Y.~Zhan, Z.~Li, and D.~Tao, ``Hl-net: Heterophily learning network for scene graph generation,'' in \emph{IEEE Conf. Comput. Vis. Pattern Recog.}, 2022, pp. 19\,454--19\,463.

\bibitem{liu2021fully}
H.~Liu, N.~Yan, M.~S. Mortazavi, and B.~Bhanu, ``Fully convolutional scene graph generation,'' in \emph{IEEE Conf. Comput. Vis. Pattern Recog.}, 2021, pp. 11\,546--11\,556.

\bibitem{li2024leveraging}
J.~Li, Y.~Wang, X.~Guo, R.~Yang, and W.~Li, ``Leveraging predicate and triplet learning for scene graph generation,'' in \emph{IEEE Conf. Comput. Vis. Pattern Recog.}, 2024, pp. 28\,369--28\,379.

\bibitem{van2008visualizing}
L.~Van~der Maaten and G.~Hinton, ``Visualizing data using t-sne.'' \emph{Journal of machine learning research}, vol.~9, no.~11, 2008.

\end{thebibliography}
}

\end{document}